\documentclass[11pt]{article}

\usepackage[preprint]{acl}

\usepackage{times}
\usepackage{latexsym}

\usepackage[T1]{fontenc}

\usepackage[utf8]{inputenc}

\usepackage{microtype}
\usepackage{enumitem}

\usepackage{inconsolata}

\usepackage{graphicx}

\renewcommand{\thefootnote}{\fnsymbol{footnote}}
\title{StoryCoder: Narrative Reformulation for Structured Reasoning in LLM Code Generation}

\author{
  Geonhui Jang$^1$, Dongyoon Han$^2$\thanks{Corresponding authors.}, YoungJoon Yoo$^1$\footnotemark[\value{footnote}] \\
  $^1$Dept. of Artificial Intelligence, Chung-Ang University \quad $^2$NAVER AI Lab \\
  \texttt{\{csleivear1, yjyoo3312\}@cau.ac.kr \quad dongyoon.han@navercorp.com}
}

\begin{document}
\maketitle

\begin{abstract}
Effective code generation requires both model capability and a problem representation that carefully structures how models reason and plan. Existing approaches augment reasoning steps or inject specific structure into how models think, but leave scattered problem conditions unchanged. Inspired by the way humans organize fragmented information into coherent explanations, we propose \mbox{\textsc{StoryCoder}}, a narrative reformulation framework that transforms code generation questions into coherent natural language narratives, providing richer contextual structure than simple rephrasings. Each narrative consists of three components: a task overview, constraints, and example test cases, guided by the selected algorithm and genre. Experiments across 11 models on HumanEval, LiveCodeBench, and CodeForces demonstrate consistent improvements, with an average gain of $18.7\%$ in zero-shot pass@10. Beyond accuracy, our analyses reveal that narrative reformulation guides models toward correct algorithmic strategies, reduces implementation errors, and induces a more modular code structure. The analyses further show that these benefits depend on narrative coherence and genre alignment, suggesting that structured problem representation is important for code generation regardless of model scale or architecture. Our code is available \href{https://github.com/gu-ni/StoryCoder}{here}.
\end{abstract}
\section{Introduction}
\label{sec:introduction}

Problem-solving ultimately depends on how clearly the information is conveyed and understood. Effective problem solving may require both the solver’s capability to interpret information and the way the problem is framed~\citep{human:Vessey, human:Kelton}. However, in practice, task descriptions are incomplete or ambiguous, forcing solvers to infer missing details from context. These gaps are especially challenging in complex tasks that demand contextual understanding or multi-step reasoning. Large language models (LLMs) face the same difficulty: their performance depends not only on internal reasoning but also on how effectively the task is specified and interpreted~\citep{llms-get-lost}. In this work, we investigate how to improve the delivery and interpretation of information in LLMs, focusing on code generation tasks. Programming tasks are particularly suitable for this study: they are built on logically distinct structures, and their solutions can be explicitly validated using test cases, making them ideal for examining how problem representation affects reasoning~\citep{plansearch, sfs}.

\begin{figure*}[t]
    \centering
    \includegraphics[width=.98\textwidth]{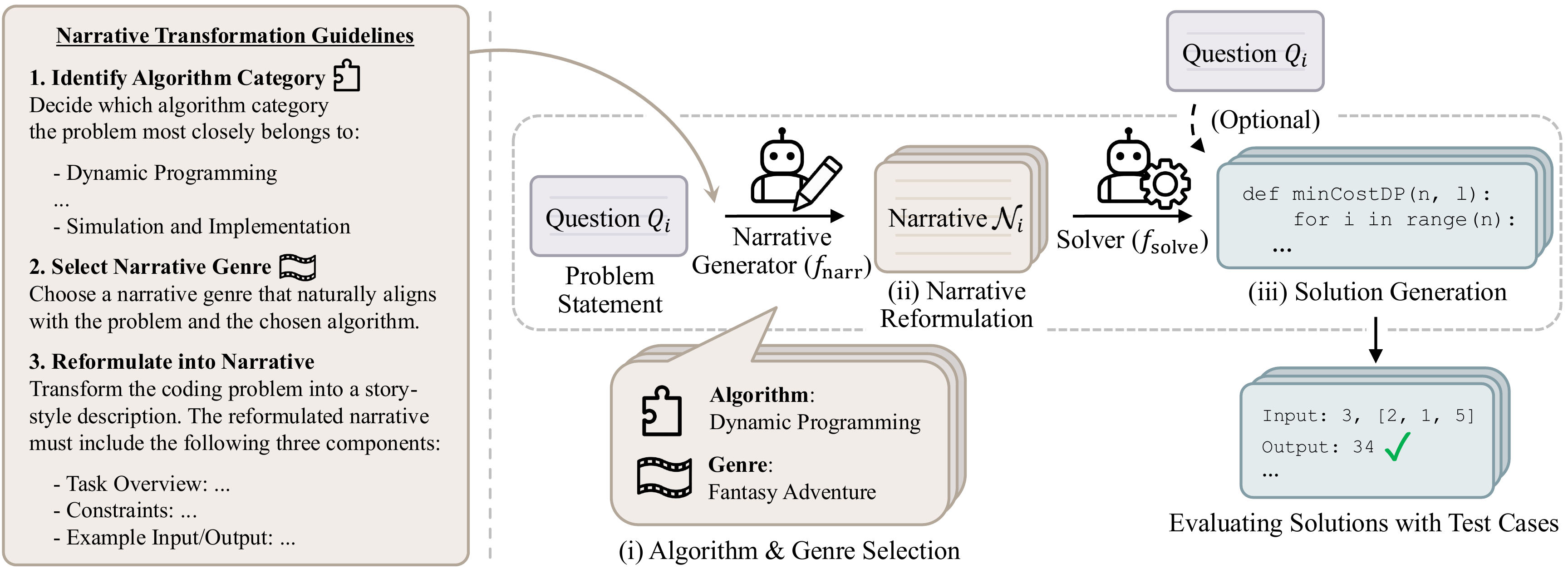}
    \caption{\textbf{Overview of \mbox{\textsc{StoryCoder}} framework.} Given a question $Q_i$, (i) model first identifies an algorithmic category and selects a narrative genre, then (ii) reformulates the problem into a structured narrative $\mathcal{N}_i$ consisting of task overview, constraints, and example input/output, and (iii) passes the narrative (optionally concatenated with $Q_i$) to a solver model to generate code solutions, which are then verified with test cases.}
    \label{fig:teaser}
\end{figure*}

We introduce \mbox{\textsc{StoryCoder}}, a narrative-based prompting method that transforms short, instruction-like problem statements into coherent natural language. This design is based on the findings of cognitive science that humans comprehend and reason more effectively when organizing fragmented conditions into coherent mental models using analogical structures~\citep{human:johnson1983mental, human:gentner1983structure, human:holyoak2021analogical}. In this framework, models identify the appropriate algorithm that will form the logic of the code, align it with a suitable narrative genre, and reformulate it into a story with three sections: task overview, constraints, and example input/output. (Figure~\ref{fig:teaser}) By connecting fragmented conditions into coherent descriptions, narratives help LLMs capture context and follow more structured reasoning. We evaluate \textsc{StoryCoder} on three benchmarks across 11 closed- and open-source models. Extensive experiments demonstrate consistent performance gains across diverse models and benchmarks.

Beyond overall accuracy, we find that narrative prompts substantially increase the likelihood of selecting the correct algorithms and reduce implementation errors. In contrast, when narratives are expressed in mismatched genres, performance drops significantly. This suggests that LLMs are sensitive to narrative structure and benefit from genre alignment in tasks such as code generation. These observations support our hypothesis that narratives are a natural and effective tool to encourage integrative reasoning.

Our key contributions are as follows:
\begin{itemize}
    \item We propose \mbox{\textsc{StoryCoder}}, a narrative-based prompting method for code generation that reformulates fragmented prompts into coherent descriptions.
    \item We demonstrate consistent empirical improvements across diverse models and benchmarks, achieving a $18.7\%$ average gain in zero-shot pass@10 accuracy.
    \item We provide quantitative analyses showing that narrative reformulation guides LLMs toward correct algorithmic strategies, reduces implementation errors, and induces more modular code structure.
\end{itemize}

\section{Related Work}
\label{sec:related_work}

\textbf{Structured prompt engineering for code generation.} 
Recent work in code generation has increasingly focused on structuring prompts to improve the reasoning process of LLMs. Some approaches introduce explicit intermediate steps, such as control structures or modular subcomponents, to guide program synthesis more reliably~\citep{scot, codechain, codecot}. Recent work demonstrates that language-based input–output patterns can enable structured and verifiable reasoning in code generation~\citep{codeio}. Related studies further investigate how narrative or prose-style problem descriptions affect requirement extraction, using narrative framing as a condition to be evaluated in code generation~\citep{pecc}. While many previous approaches present structured reasoning through explicit intermediate steps or modular decomposition, our method \mbox{\textsc{StoryCoder}} provides a structured reasoning trajectory to the model through narrative-based prompt design.

\noindent\textbf{Prompt reformulation and test-time reasoning in LLMs.} 
Another line of research examines how rephrasing prompts at test time affects the way LLMs reason on a task. Chain-of-thought prompting, especially when combined with self-consistency, helps models explore diverse reasoning paths and improves robustness through output aggregation~\citep{cot, sc}. Beyond explicit reasoning steps, rephrasing task instructions influences how the model interprets the problem~\citep{learning-to-paraphrase-for-alignment, scop}. In the context of multiple-choice reasoning tasks, narrative-based prompting has been explored as a structured reasoning framework to support the selection of answers among predefined choices~\citep{sot}. Building on these insights, our method reformulates prompts into coherent task-grounded narratives that integrate conditions, intent, and examples within a unified structure for code generation.

\begin{figure*}[t]
    \centering
    \includegraphics[width=.98\textwidth]{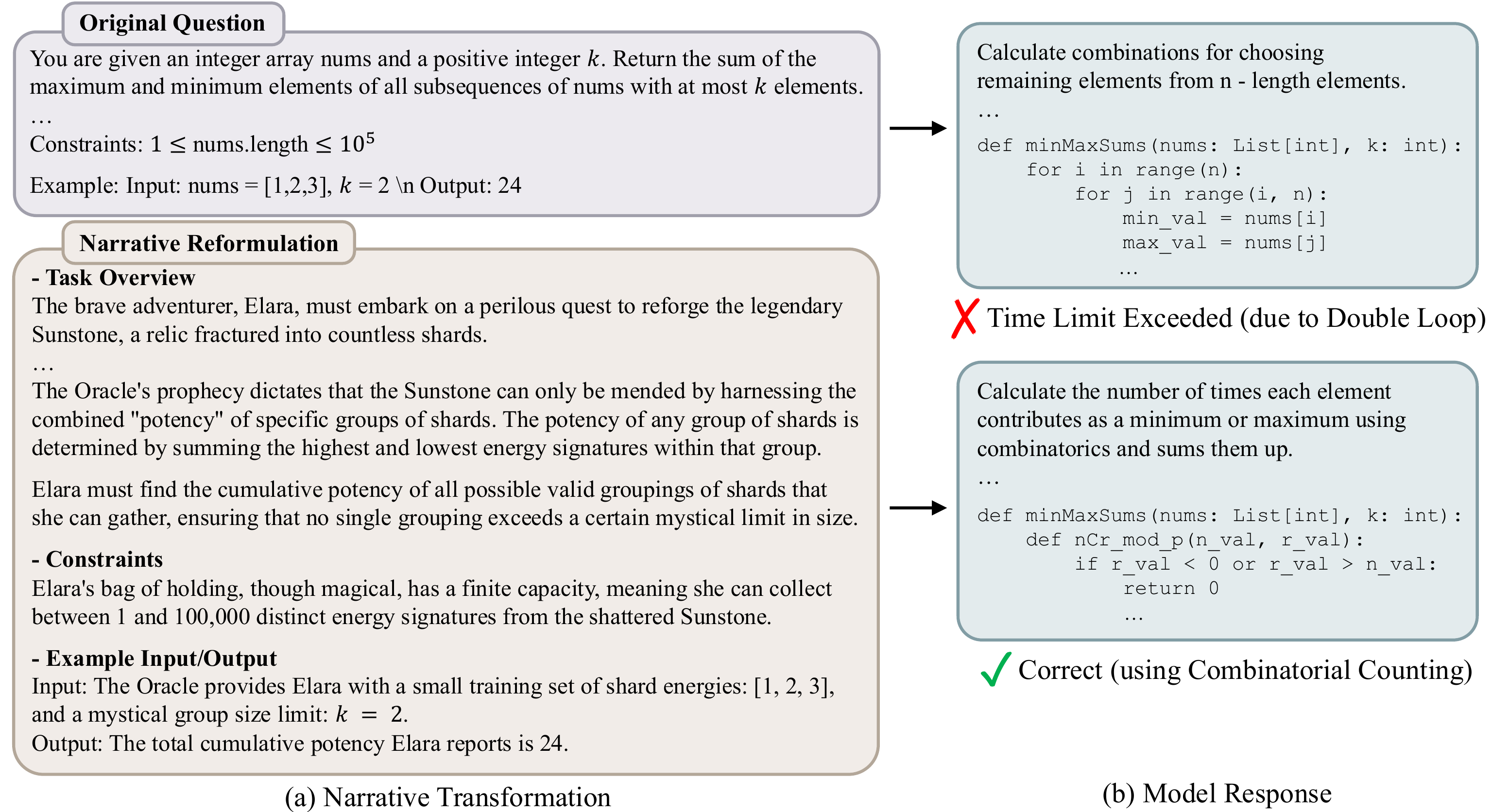}
    \caption{\textbf{Example of narrative reformulation.} The narrative representation bridges problem description and model reasoning, guiding the model from inefficient non-optimal solutions toward algorithmic strategies.}
    \label{fig:main_example}
\end{figure*}
\section{\mbox{\textsc{StoryCoder}}: Reformulating question into narratives}
\label{sec:method}

\subsection{Conventional baselines for code generation}
\label{sec:3.1}

Early evaluations of code generation relied on similarity-based metrics such as CodeBLEU, which were insufficient for assessing functional correctness. Recent benchmarks adopt execution-based evaluation using the pass@$k$ metric~\citep{humaneval}. Under this setting, we consider the following baselines: Repeated Sampling, Paraphrasing, and Chain-of-Thought (CoT) prompting.
\vspace{-.3em}
\begin{itemize}[leftmargin=*, itemsep=1pt]
\item  \textbf{Repeated sampling} improves pass@$k$ by generating multiple outputs for the same input via stochastic decoding, typically with high temperature~\citep{humaneval}. Increasing the number of candidates raises the likelihood that at least one solution is correct.
\item \textbf{Chain-of-thought (CoT)} prompts the model to generate intermediate reasoning steps prior to producing code~\citep{cot, codecot}. By making the reasoning process explicit, CoT encourages stepwise planning.
\item \textbf{Structured chain-of-thought (SCoT)} extends CoT for code generation by incorporating program structures (sequence, branch, and loop) into the intermediate reasoning steps~\citep{scot}. It guides the model to think from the perspective of the source code before generating the output.
\end{itemize}

\subsection{Narrative reformulation for deeper comprehension}
\label{sec:3.2}

The approaches discussed in Section~\ref{sec:3.1} are useful, but they have a limitation: they either expand outputs without changing the input, or introduce reasoning steps that do not always reflect how models actually process the problem~\citep{dont_always_say}. Research in cognitive science suggests that more effective reasoning emerges when fragmented conditions are organized into a unified relational structure~\citep{human:gentner1983structure}. Inspired by this, we propose a narrative reformulation framework that reorganizes task representation, encouraging the model to form a more coherent and semantically grounded understanding of the input.

We construct a framework that reformulates code generation questions into a narrative format in three stages as shown in Figure~\ref{fig:teaser}: (i) for the $i$-th question $Q_i$, choosing an appropriate algorithmic category $a_i$ and a narrative genre $g_i$; (ii) rewriting $Q_i$ as a structured narrative $\mathcal{N}_i$ with three parts: task overview, constraints, and example input/output; and (iii) solving the task using $\mathcal{N}_i$. Here, the algorithm $a_i$ denotes the algorithmic category that the model judges to be the most appropriate for the given problem, chosen from the eight predefined categories in Figure~\ref{fig:appendix_narrative_transformation_guidelines}. The genre $g_i$ denotes the narrative style, selected freely by the model to align with the problem and the chosen algorithm $a_i$. We generate $N$ narrative variants for each question $Q_i$ and denote the index of each reformulation variant by $j \in \{1, \dots, N\}$.  Note that generating these narrative variants differs from simply drawing multiple solutions, as each narrative provides a distinct perspective and plot that broadens the model’s representational space for interpreting and reasoning about the task. Formally, the overall pipeline can be described as follows:
\begin{equation}
\begin{aligned}
\{a_i^j, \; g_i^j, \; \mathcal{N}_i^j\}_{j=1}^N
&= f_{\text{narr}}(Q_i) \\
\text{Ans}(\mathcal{N}_i^j)
&= f_{\text{solve}}(\mathcal{N}_i^j),
\end{aligned}
\label{eq:narr-solve}
\end{equation}
where $f_{\text{narr}}$ is the generator model that formulates narratives and $f_{\text{solve}}$ is the solver model that generates code solutions $\text{Ans}(\cdot)$. We refer to the case where $f_{\text{narr}}$ and $f_{\text{solve}}$ are the same model as the self-solving setting, and the case where they differ as the cross-model setting. Note that $\mathcal{N}_i^j \sim P(\,\cdot \mid a_i^j, g_i^j)$, where $\mathcal{N}_i^j$ denotes the $j$-th narrative variant of the question $Q_i$, and $P$ denotes the conditional distribution of narratives given the algorithmic category $a_i^j$ and genre $g_i^j$ selected by $f_{\text{narr}}$.

In stage (ii), unlike the prior work~\citep{sot}, we carefully design the narrative components for programming tasks, where precise format and formal constraints are required. To ensure that the reformulated prompt preserves both narrative coherence and computational strictness, we divide a narrative $\mathcal{N}_i^j$ into three parts:

\vspace{-0.3em}
\begin{itemize}[leftmargin=*, itemsep=1pt]
    \item \textbf{Task Overview} $(\text{TO}_i^j)$: presents the coding objective within a narrative frame, integrating scattered conditions into a coherent system that guides comprehension and reasoning.
    \item \textbf{Constraints} $(\text{C}_i^j)$: reframes input ranges, time limits, and rules as natural restrictions in the story, allowing the model to internalize constraints within the narrative space.
    \item \textbf{Example Input/Output} $(\text{E}_i^j)$: integrates sample test cases into contextual scenarios, aligning input/output examples with the story structure, so that formal coding task requirements are preserved within the narrative space.
\end{itemize}
This three-part structure is motivated by a principle observed in both cognitive science and recent analyses of LLM behavior: effective reasoning depends on forming a coherent and specified problem representation before solution generation. Cognitive science research on mental models suggests that humans reason by constructing structured representations that capture the essential relations of a situation within a unified mental representation~\citep{human:johnson1983mental}. Similarly, recent studies on LLM prompting have shown that underspecified or fragmented problem descriptions can lead to degradation in model performance~\citep{whatpromptsdontsay}. \mbox{\textsc{StoryCoder}} builds on this observation by organizing task overview, constraints, and examples into a single narrative, with the goal of encouraging the model to form a consistent high-level problem representation prior to code generation.

Together, these three components of the narrative $\mathcal{N}_i^j = \{\text{TO}_i^j, \; \text{C}_i^j, \; \text{E}_i^j\}$ standardize the reformulation process, ensuring that all essential details of the original question of code generation are preserved while also allowing cognitive principles to be naturally integrated into the narrative structure. Transformation examples are provided in Figure~\ref{fig:main_example} and Appendix~\ref{appendix:narrative_transformation_examples}.
\section{Experiments}
\label{sec:experiments}
\subsection{Experimental settings}
\begin{table*}[t]
\caption{\textbf{Pass@10 performance} on three benchmarks. The upper part of the table reports closed-source models, while the lower part reports open-source models. We evaluate \mbox{\textsc{StoryCoder}} against representative prompting baselines across 11 models to assess its generality. Our method consistently outperforms repeated sampling (RS), CoT, and SCoT across the benchmarks and models, especially with larger gains on challenging benchmarks.}
\centering
\small
\tabcolsep=.5em
\begin{tabular}{l|cccc|cccc|cccc}
\toprule
 & \multicolumn{4}{c|}{HumanEval} & \multicolumn{4}{c|}{LiveCodeBench} & \multicolumn{4}{c}{CodeForces} \\
Model & RS & CoT & SCoT & Narr. & RS & CoT & SCoT & Narr. & RS & CoT & SCoT & Narr.  \\
\midrule
Gemini-2.5-Flash & \textbf{96.19} & 95.19 & 95.10 & \textbf{96.19} & 53.14 & 50.63 & 50.86 & \textbf{57.14} & 60.00 & 53.19 & 52.13 & \textbf{67.55} \\
GPT-4.1-mini     & \textbf{96.19} & 94.29 & 95.24 & 94.29 & 50.29 & 52.57 & 49.71 & \textbf{56.57} & 38.87 & 43.02 & 46.41 & \textbf{50.57} \\
Claude-3.5-Haiku & 85.71 & 83.81 & 78.10 & \textbf{94.29} & 33.71 & 29.14 & 26.29 & \textbf{38.29} & 47.17 & 19.62 & 24.15 & \textbf{50.95} \\
\rowcolor{gray!15}
Average          & 92.70 & 91.10 & 89.48 & \textbf{94.92} & 45.71 & 44.11 & 42.29 & \textbf{50.67} & 48.68 & 38.61 & 40.90 & \textbf{56.36} \\
\midrule
DSCoder 6.7B       & 82.86 & 83.81 & 84.31 & \textbf{90.48} & 22.29 & 22.86 & 23.43 & \textbf{27.43} & 14.34 & 11.69 & 11.84 & \textbf{19.62} \\
DSCoder-V2-Lite    & 78.10 & 80.95 & 85.29 & \textbf{93.33} & 28.57 & 29.14 & 26.29 & \textbf{34.29} & 26.04 & 26.28 & 23.88 & \textbf{33.96} \\
Llama-3.1 8B       & 79.05 & 76.19 & 75.49 & \textbf{81.90} & 21.14 & 24.57 & 22.86 & \textbf{27.43} &  9.44 & 10.92 &  8.83 & \textbf{20.38} \\
Gemma-2 9B         & 63.81 & 62.86 & 60.78 & \textbf{82.86} & 20.00 & 19.43 & 20.57 & \textbf{26.29} & 12.83 & 10.74 & 12.62 & \textbf{22.26} \\
Gemma-2 27B        & 76.19 & 80.95 & 77.45 & \textbf{87.62} & 27.43 & 25.71 & 25.71 & \textbf{34.29} & 23.77 & 22.38 & 20.64 & \textbf{32.07} \\
Qwen-2.5-Coder 7B  & 89.52 & 92.38 & 92.16 & \textbf{93.33} & 26.86 & 29.71 & 26.86 & \textbf{33.14} & 19.62 & 20.58 & 18.55 & \textbf{27.92} \\
Qwen-2.5-Coder 32B & 92.38 & 90.48 & \textbf{95.10} & 94.29 & 30.86 & 34.29 & 36.00 & \textbf{40.00} & 15.47 & 24.98 & 24.72 & \textbf{28.68} \\
Mistral-Small 24B  & 88.57 & 90.48 & 90.20 & \textbf{94.29} & 33.71 & \textbf{34.86} & 33.71 & \textbf{34.86} & 30.18 & 27.01 & 33.00 & \textbf{43.77} \\
\rowcolor{gray!15}
Average            & 81.31 & 82.26 & 82.60 & \textbf{89.76} & 26.36 & 27.57 & 26.93 & \textbf{32.22} & 18.96 & 19.32 & 19.26 & \textbf{28.58} \\
\bottomrule
\end{tabular}
\label{tab:main_scores}
\end{table*}

\noindent\textbf{Models.} We evaluate a total of 11 models of varying sizes. Among open-source models, we use the instruction-tuned versions of Deepseek-Coder 6.7B~\citep{deepseekcoder}, Deepseek-Coder-V2-Lite~\citep{deepseekcoderv2}, Llama-3.1 8B~\citep{llama3}, Gemma-2 9B and 27B~\citep{gemma2}, Qwen-2.5-Coder 7B and 32B~\citep{qwen2.5coder}, and Mistral-Small 24B~\citep{mistralsmall}. For readability, we omit `Instruct' from all the model names below. For closed-source models, we include Claude-3.5-Haiku~\citep{claude3.5haiku}, Gemini-2.5-Flash~\citep{gemini2.5}, and GPT-4.1-mini~\citep{gpt4.1mini}. The narrative and code are generated with a temperature of 1.0 and 0.2.

\noindent\textbf{Dataset.} We evaluate on three benchmarks: HumanEval~\citep{humaneval}, a dataset with function signatures and docstrings; LiveCodeBench~\citep{livecodebench}, a large-scale dataset covering various programming problems; and CodeForces~\citep{codeforces}, real-world algorithmic problems from competitive programming. For HumanEval, we exclude questions that are invalid or contain incorrect sample input/output, resulting in a filtered set of 105 questions. For LiveCodeBench, we use the 175 questions from release-v6, sourced from AtCoder~\citep{atcoder} or LeetCode~\citep{leetcode}. For CodeForces, we apply filtering based on question length and difficulty, yielding a final set of 265 questions. For detailed filtering criteria and additional results, see the Appendix~\ref{appendix:dataset_filtering}.

\noindent\textbf{Metric and evaluation.} We report the results using the pass@$k$ metric. The pass@$k$ metric measures the probability that at least one correct solution is found among $k$ sampled outputs. Following HumanEval, we consider a generated solution to be correct only when it passes all test cases (see Appendix~\ref{appendix:evaluation_metric}.)

In our experiments, we set $N=5$ narrative variants per question. For the narrative setting in the main paper, we aggregate 10 total responses per question: five narrative-only variants $\{\mathcal{N}_i^j\}_{j=1}^5$ and five narrative concatenated with the original question $\{\mathcal{N}_i^j, Q_i\}_{j=1}^5$. Both forms follow the same reformulation pipeline, with the original question appended as additional input in the concatenated variant. To ensure fairness, the repeated sampling baseline generates 10 samples per question, matching the total number of samples used in the narrative setting. The pass@5 results for each individual set of five variants are reported in Appendix~\ref{appendix:comprehensive_results}.

\begin{table}[t]
\normalsize
\caption{\textbf{Proportion of valid narratives} where the model $f_{\text{narr}}$ follows the transformation guidelines properly. For the DeepSeek (DS) and Qwen families, we use their base instruction-tuned versions (DeepSeek-V2-Lite-Chat~\citep{deepseekcoderv2}, Qwen2.5-7B/32B-Instruct~\citep{qwen2.5}) rather than coding-specialized variants, as narrative transformation is closer to a basic instruction-following task.}
\centering
\small
\setlength{\tabcolsep}{3pt}
\begin{tabular}{lccccccc}
\toprule
& DS & \multicolumn{2}{c}{Gemma} & Llama & Mistral & \multicolumn{2}{c}{Qwen} \\
\cmidrule(lr){2-2}\cmidrule(lr){3-4}\cmidrule(lr){5-5}\cmidrule(lr){6-6}\cmidrule(lr){7-8}
Model & V2 & 9B & 27B & 8B & 24B & 7B & 32B \\
\midrule
Valid (\%) & 68.1 & 51.7 & 96.0 & 36.7 & 86.9 & 37.8 & 76.6 \\
\bottomrule
\end{tabular}
\label{tab:open_sourced_models_valid_proportion}
\end{table}

\begin{table*}[t]
\caption{\textbf{\texorpdfstring{\boldmath Pass@$k$}{Pass@k} performance of open-source models} for self-solving scenarios. N-Q, N-M, and N-G correspond to $f_{\text{narr}}$ using Qwen2.5 32B Instruct, Mistral-Small 24B Instruct, and Gemma 2 27B Instruct, respectively. Repeated sampling (RS) is evaluated with pass@10. Narrative-based scores are computed as pass@$k$ with $k = n \in [8, 10]$, using the maximum number of valid narratives per question, and the Used Samples Ratio denotes the proportion of samples used for scoring. Narratives generated by open-source models still improve performance.}
\centering
\small
\tabcolsep=0.5em
\begin{tabular}{l|cccc|cccc|cccc}
\toprule
 & \multicolumn{4}{c|}{HumanEval} 
 & \multicolumn{4}{c|}{LiveCodeBench} 
 & \multicolumn{4}{c}{CodeForces} \\
                   & RS & N-Q & N-M & N-G 
                   & RS & N-Q & N-M & N-G
                   & RS & N-Q & N-M & N-G \\
Used Samples Ratio & 100.0 & 64.42 & 97.14 & 95.24 
                   & 100.0 & 68.00 & 79.77 & 100.0
                   & 100.0 & 65.49 & 86.5 & 98.9 \\
\midrule
DSCoder 6.7B       & 82.86 & \textbf{92.54} & 86.27 & 87.0 
                   & 22.29 & 19.33 & \textbf{23.19} & 22.86 
                   & 13.12 & 12.34 & \textbf{13.14} & 12.92 \\
DSCoder V2 Lite    & 78.10 & 91.04 & \textbf{92.16} & 90.0 
                   & 28.57 & 28.57 & \textbf{31.88} & 29.71 
                   & 25.16 & 30.00 & \textbf{32.25} & 28.17 \\
Llama 3.1 8B       & 79.05 & \textbf{85.07} & 81.37 & 84.0
                   & 21.14 & 22.69 & \textbf{25.36} & 24.0
                   & \ 8.11 & 14.34 & \textbf{14.56} & 14.20 \\
Gemma 2 9B         & 63.81 & \textbf{86.57} & 85.29 & 85.0
                   & 20.00 & 19.33 & 22.46 & \textbf{22.86}
                   & 11.69 & 17.23 & 17.36 & \textbf{18.36} \\
Gemma 2 27B        & 76.19 & 88.06 & \textbf{88.24} & 84.0
                   & 27.43 & 25.21 & 26.09 & \textbf{28.57}
                   & 22.38 & 27.86 & \textbf{30.18} & 27.06 \\
Qwen 2.5 Coder 7B  & 89.52 & 94.03 & \textbf{94.12} & 90.0
                   & 26.86 & 28.57 & \textbf{33.33} & 29.71
                   & 19.30 & 21.02 & \textbf{21.82} & 21.78 \\
Qwen 2.5 Coder 32B & 92.38 & \textbf{95.52} & 94.12 & 93.0
                   & 30.86 & 32.77 & 36.96 & \textbf{40.57}
                   & 15.08 & 23.40 & 24.02 & \textbf{26.06} \\
Mistral-Small 24B  & 88.57 & \textbf{92.54} & 92.16 & 89.0
                   & 33.71 & 31.93 & \textbf{35.51} & 32.57
                   & 29.23 & 32.23 & \textbf{34.68} & 33.52 \\
\rowcolor{gray!15}
Average            & 81.31 & \textbf{90.67} & 89.22 & 87.75
                   & 26.36 & 26.05 & \textbf{29.35} & 28.86
                   & 18.01 & 22.30 & \textbf{23.50} & 22.76 \\
\bottomrule
\end{tabular}
\label{tab:open_sourced_models_scores}
\end{table*}

\subsection{Experimental results}
\label{sec:4.2}

Table \ref{tab:main_scores} presents pass@10 results on three coding benchmarks. For closed-source models, we adopt a self-solving setting where $f_{\text{narr}} = f_{\text{solve}}$, while for open-source models, we adopt a cross-model setting, i.e., the narratives are generated by Gemini-2.5-Flash ($f_{\text{narr}}$) and solved by each open-source model ($f_{\text{solve}}$). Repeated sampling relies solely on stochastic decoding rather than exploring representation diversity, CoT often produces unstructured explanations that are inappropriate for programming tasks, and SCoT introduces program-level structure but does not alter how the problem itself is represented. Meanwhile, narrative prompting consistently outperforms all baselines across all benchmarks and models, demonstrating its general effectiveness for code generation. Improvements are more pronounced on challenging benchmarks such as CodeForces and LiveCodeBench. Additionally, the pass@$k$ curves in Appendix~\ref{appendix:pass_k_curve} show that narrative prompting outperforms the baseline as $k$ increases. Additional experimental results, including evaluations on the latest models, are provided in Appendix~\ref{appendix:additional_results}.

We initially considered evaluating open-source models in a strict self-solving setting (i.e., $f_{\text{narr}} = f_{\text{solve}}$). However, as shown in Table~\ref{tab:open_sourced_models_valid_proportion}, open-source models often fail to reliably follow the required narrative format, resulting in substantial differences in valid narrative ratios. To enable a more equitable comparison, we select the top-3 open-source models with the highest valid narrative rates (Qwen 2.5 32B Instruct, Mistral-Small 24B Instruct, and Gemma 2 27B Instruct) and use them as narrative generators ($f_{\text{narr}}$). Although this setup is not strict self-solving, this relaxed configuration allows us to assess whether narrative reformulation improves performance when narratives are produced by open-source models rather than relying on a strong generator. For detailed filtering criteria and examples of invalid narratives, refer to Appendix~\ref{appendix:narrative_validity_filtering_criteria}.

In Table~\ref{tab:open_sourced_models_scores}, even in the self-solving setting of open-source models, \mbox{\textsc{StoryCoder}} consistently improves performance across all benchmarks. On HumanEval, narratives generated by Qwen 2.5 32B (N-Q) show the strongest improvements across solvers. In contrast, on both LiveCodeBench and CodeForces, Mistral-Small 24B narratives (N-M) achieve the highest performance. These results suggest that narrative effectiveness for open-source models depends on how well the generator constructs and interprets the reformulation, with the optimal choice varying by the pretrained knowledge of the generator and the complexity of the target benchmark.

\begin{table}[t]
    \centering
    \small
    \caption{
        \textbf{Pass@10 performance} with and without Example I/O. The gap between RS and \mbox{\textsc{StoryCoder}} widens when Example I/O is removed, demonstrating that narrative semantics alone are sufficient to enhance reasoning.
    }
    \resizebox{\columnwidth}{!}{%
    \setlength{\tabcolsep}{4.5pt}
    \begin{tabular}{l|cc|cc|cc}
    \toprule
     & \multicolumn{2}{c|}{HumanEval} & \multicolumn{2}{c|}{LiveCodeBench} & \multicolumn{2}{c}{CodeForces} \\
    Setting & RS & Narr. &\ RS & Narr. & RS & Narr. \\
    \midrule
    w/ I/O    & \textbf{96.19} & \textbf{96.19} &\ 53.14 & \textbf{57.14} & 60.00 & \textbf{67.55} \\
    w/o I/O  & 93.33 & \textbf{96.19} &\ 38.29 & \textbf{52.00} & 35.10 & \textbf{57.36} \\
    \bottomrule
    \end{tabular}
    }
    \label{tab:example_io_ablation}
    \vspace{-.5em}
\end{table}

\subsection{Ablation study}
\label{sec:4.3}

To isolate the contribution of narrative semantics from that of example I/O, we conduct an ablation study comparing repeated sampling~(RS) and \mbox{\textsc{StoryCoder}} with and without example I/O across all three benchmarks using Gemini-2.5-Flash. The results in Table~\ref{tab:example_io_ablation} reveal two key findings. First, both methods degrade without examples, confirming that it is a fundamental component across prompting strategies. Second, the performance drop of \mbox{\textsc{StoryCoder}} is substantially smaller than that of RS~(avg. 5.11\%p vs. 14.20\%p), demonstrating that narrative semantics alone, through reformulated task overview and constraints, are sufficient to enhance reasoning even in the absence of examples. We further note that this binding is deliberately designed: example I/O is inherent to code generation benchmarks, and its coherent integration into the narrative is itself a meaningful contributor to the gains. For additional ablations, see Appendix~\ref{appendix:ablation_study}.
\section{Discussion}
\label{sec:discussion}

In this section, we analyze how and why narrative reformulation improves code generation from multiple perspectives: whether narratives expand the solution space with algorithmic agreement (Section~\ref{sec:5.1}), error decomposition at the algorithm and implementation levels (Section~\ref{sec:5.2}), the role of narrative coherence among components (Section~\ref{sec:5.3}), whether LLMs recognize an optimal narrative space (Section~\ref{sec:5.4}), and code-level properties of generated solutions (Section~\ref{sec:5.5}). For Sections~\ref{sec:5.1} and~\ref{sec:5.2}, given a code solution generated by the solver model $f_{\text{solve}}$, we query another model instance $f_{\text{alg}}$ to identify the algorithm underlying each generated solution. This procedure is similar to back-translation in machine translation~\citep{machine_translation, plansearch}. All experiments are conducted with Gemini-2.5-Flash.

\begin{figure}[t]
  \centering
  \includegraphics[width=0.75\columnwidth]{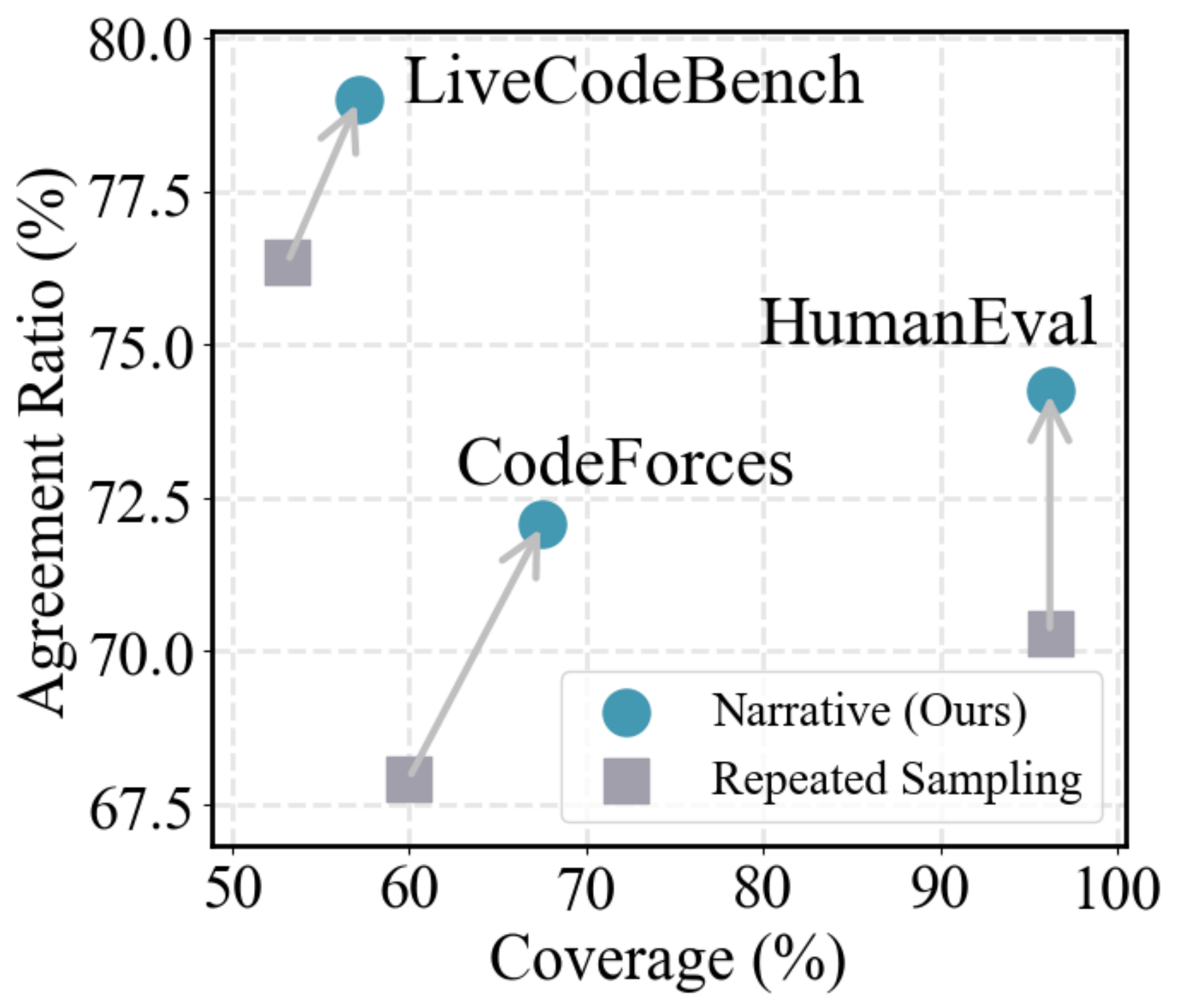}
  \vspace{-0.5em}
  \caption{\textbf{Effect of narrative reformulation.} The x-axis denotes coverage (pass@10), and the y-axis shows the agreement ratio, the proportion of correct solutions consistent with the initial chosen algorithm, $a_i$. Narrative reformulation simultaneously achieves broader coverage and higher algorithmic fidelity.}
  \label{fig:agreement}
\end{figure}

\begin{figure}[t]
    \centering
    \includegraphics[width=1.0\columnwidth]{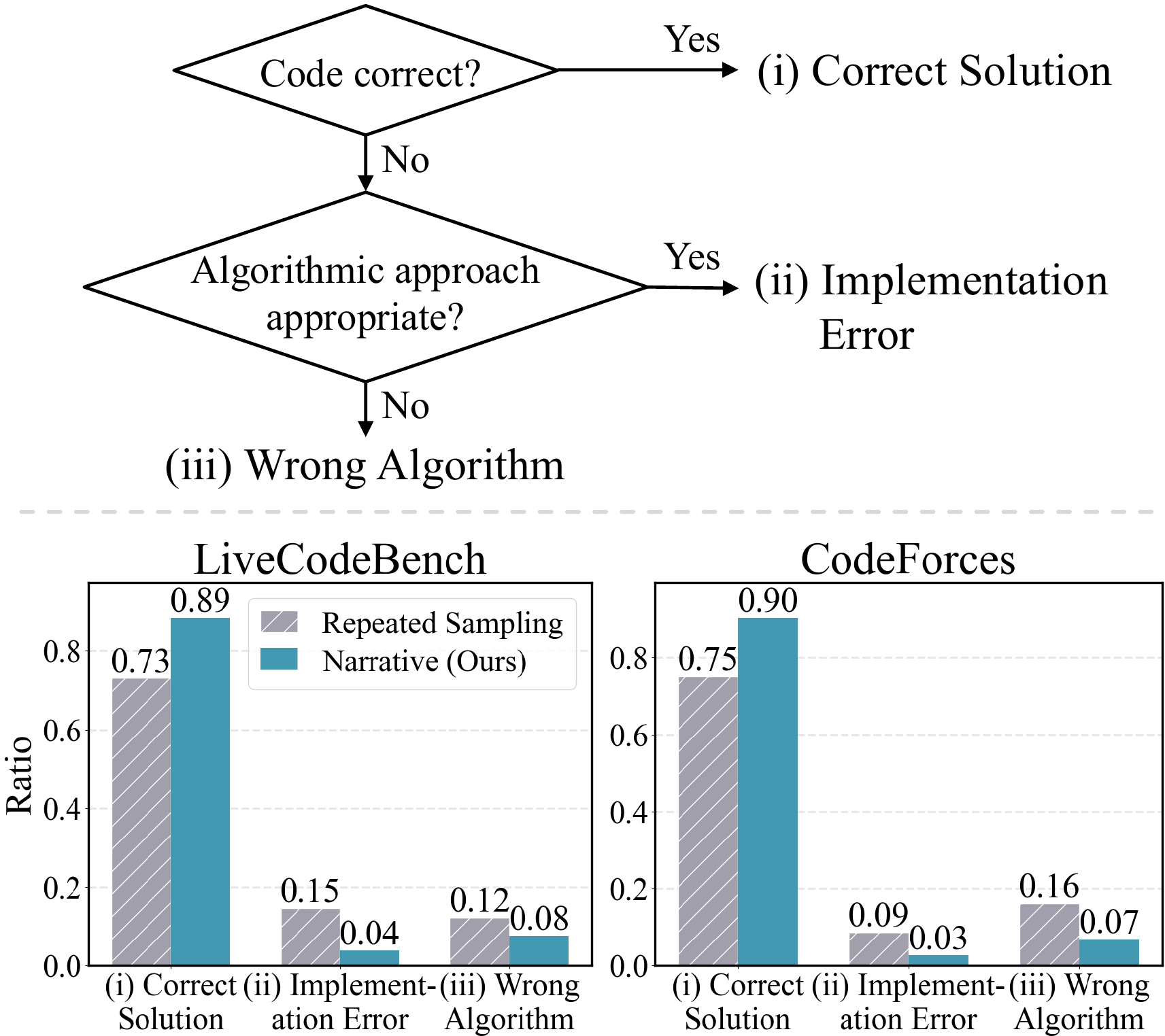}
    \vspace{-1.5em}
    \caption{\textbf{Decomposition of model outputs} into (i) correct solution, (ii) implementation errors, and (iii) incorrect algorithm choice. (\textit{Top}) Flowchart for categorizing model outputs. (\textit{Bottom}) Ratio of each category under repeated sampling vs. \mbox{\textsc{StoryCoder}}. Narrative reformulation steers models toward correct algorithmic strategies and implementations.}
    \label{fig:error_decomposition}
\end{figure}

\subsection{Coverage and algorithmic agreement}
\label{sec:5.1}

We use $f_{\text{alg}}$ to extract back-translated algorithms from the outputs of $f_{\text{solve}}$ and compare them with $a_i^j$, the algorithms initially predicted by $f_{\text{narr}}$ and incorporated into the narratives. Formally, we define the agreement ratio as the fraction of correct solutions whose back-translated algorithms match $a_i^j$. We also define coverage as the proportion of problems for which at least one correct solution is generated, which is equivalent to pass@$k$ when $k$ equals the number of generated solutions.

In Figure~\ref{fig:agreement}, coverage increases on challenging benchmarks such as LiveCodeBench and CodeForces, indicating that narratives expand the solution space, especially for harder tasks. The agreement ratio increases across all benchmarks, demonstrating that the initially selected algorithm is faithfully reflected in the generated solutions. Together, these results show that \emph{narratives enable broader coverage and strengthen algorithmic agreement}.

\subsection{Decomposing narrative contributions}
\label{sec:5.2}

Solutions to code generation tasks can be decomposed into algorithms or sketch ideas, where selecting the right algorithm is necessary to arrive at a correct solution~\citep{plansearch}. To analyze how models follow this process, we categorize their outputs into three outcomes as shown in Figure~\ref{fig:error_decomposition}: (i) correct solutions, (ii) incorrect responses where the chosen algorithm is appropriate but the implementation leads to an error, and (iii) incorrect responses by selecting the wrong algorithm.

For steps (ii) and (iii) of the categorization, we automatically extract a golden algorithm using $f_{\text{alg}}$. Specifically, we take all generated code solutions confirmed to be correct, from either the original or narrative problems, and query $f_{\text{alg}}$ to back-translate the algorithm. We then determine the golden algorithm $a_i^*$ by majority voting among the candidates:
\begin{equation*}
\begin{aligned}
a_i^*
&= \text{MajorityVote}\Big(
\Big\{
f_{\text{alg}}\!\left(\text{Ans}(X_i)\right)
\;\Big|\;
\\
&\quad
X_i \in Q_i \cup \{\mathcal{N}_i^j\}_{j=1}^N,\;
\text{Ans}(X_i)\ \text{is correct}
\Big\}
\Big).
\end{aligned}
\label{eq:gold-algo}
\end{equation*}
The $a_i^*$ is used to evaluate whether incorrect solutions nevertheless adopt the correct algorithm, with such cases classified as (ii) implementation errors (e.g., incorrect loop bounds). To better isolate the effect of representation change, we exclude trivial cases where generations for both original and narrative are either all correct or all incorrect.

As shown in Figure~\ref{fig:error_decomposition}, across both benchmarks, narratives increase the proportion of correct solutions while reducing both implementation errors and incorrect algorithm choices. This suggests that \emph{narrative reformulation improves procedural reasoning at both the algorithm selection and implementation levels.}

\subsection{Component coherence matters}
\label{sec:5.3}

Coherent integration of information components facilitates comprehension more effectively than disparate sources~\citep{human:chandler1991cognitive}. Building on this, we design experiments that permute narrative components across variants to examine whether coherent integration contributes to the performance gains, allowing us to distinguish between the informational benefits of narratives and the additional gains from the integration.

\begin{figure}[t]
    \centering
    \includegraphics[width=1\columnwidth]{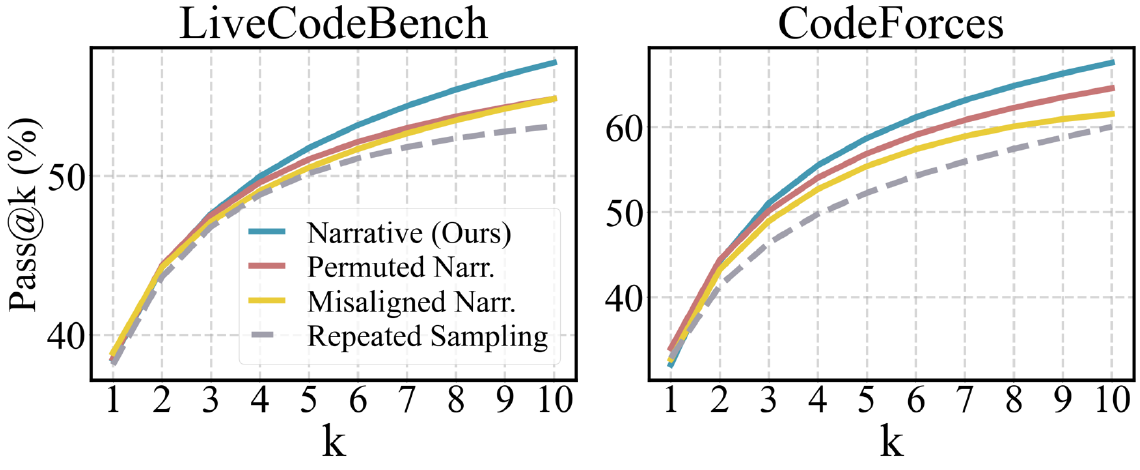}
    \vspace{-1.5em}
    \caption{\textbf{Comparison of \texorpdfstring{\boldmath pass@$k$}{pass@k} curves across different prompt settings}. Permuted narratives (components mixed across variants) outperform original prompts but remain below complete narratives, indicating the importance of coherence (Section~\ref{sec:5.3}). Misaligned narratives (genres forced from incongruent sets) degrade performance, showing that proper representation contributes to effective problem solving (Section~\ref{sec:5.4}).}
    \label{fig:permuted_and_misaligned}
\end{figure}

Recall that each narrative reformulation $\mathcal{N}_i^j$, the $j$-th variant of $Q_i$, consists of three parts: $\mathcal{N}_i^j = \{\text{TO}_i^j, \; \text{C}_i^j, \; \text{E}_i^j\}$. In the permuted setting, we construct a new narrative $\widetilde{\mathcal{N}}_i^{j_1, j_2, j_3}$ by sampling these components from different variants:
\begin{equation*}
\begin{aligned}
\widetilde{\mathcal{N}}_i^{j_1, j_2, j_3}
&= \{\text{TO}_i^{j_1}, \; \text{C}_i^{j_2}, \; \text{E}_i^{j_3}\}, \\
& \text{where } j_1 \neq j_2,\; j_1 \neq j_3,\; \text{and } j_2 \neq j_3.
\end{aligned}
\end{equation*}
We compare three conditions: \textbf{Original} ($Q_i$), without narrative reformulation; \textbf{Complete Narrative} ($\mathcal{N}_i^j$), all components from the same variant $j$; and \textbf{Permuted Narrative} ($\widetilde{\mathcal{N}}_i^{j_1, j_2, j_3}$), each component drawn from a distinct variant. Surprisingly, permuted narratives still outperform the original prompts across all $k$ as shown in Figure~\ref{fig:permuted_and_misaligned}. Narrative reformulation provides informational benefits even when components are drawn from different variants. However, the permuted narratives fall short of the complete narratives, suggesting that \emph{narrative components are most effective when they form a coherent whole, with the full gains obtained through a unified structure.}

\begin{figure}[t]
    \centering
    \small
    \begin{subfigure}{\columnwidth}
        \centering
        \setlength{\tabcolsep}{0.1pt}
        \begin{tabular}{c@{\hspace{3pt}}|@{\hspace{5pt}}l@{\hspace{-3pt}}c}
        \toprule
        \textbf{Model} & \textbf{Narrative Genre} & \textbf{Ratio (\%)} \\
        \midrule
        \multirow{3}{*}{Gemini} 
         & ``Fantasy Adventure''    & 12.5 \\
         & ``Sci-Fi / Exploration'' & 7.0  \\
         & ``Fantasy / Quest''      & 6.9  \\
        \midrule
        \multirow{3}{*}{ChatGPT} 
         & ``Fantasy Adventure''    & 23.7 \\
         & ``Fantasy Quest''        & 17.1 \\
         & ``Epic Fantasy Quest''   & 14.8 \\
        \midrule
        \multirow{4}{*}{Claude} 
         & ``Mathematical Mystery'' & 6.3 \\
         & \makecell[l]{``Mathematical Mystery\\Adventure''} & 5.5 \\
         & \makecell[l]{``Strategic Puzzle Adventure''} & 4.0 \\
        \bottomrule
        \end{tabular}
        \caption{Top three narrative genres selected by each model.}
        \label{fig:permuted_table}
    \end{subfigure}
    \begin{subfigure}{\columnwidth}
        \centering
        \vspace{.5em}
        \includegraphics[width=.92\linewidth]{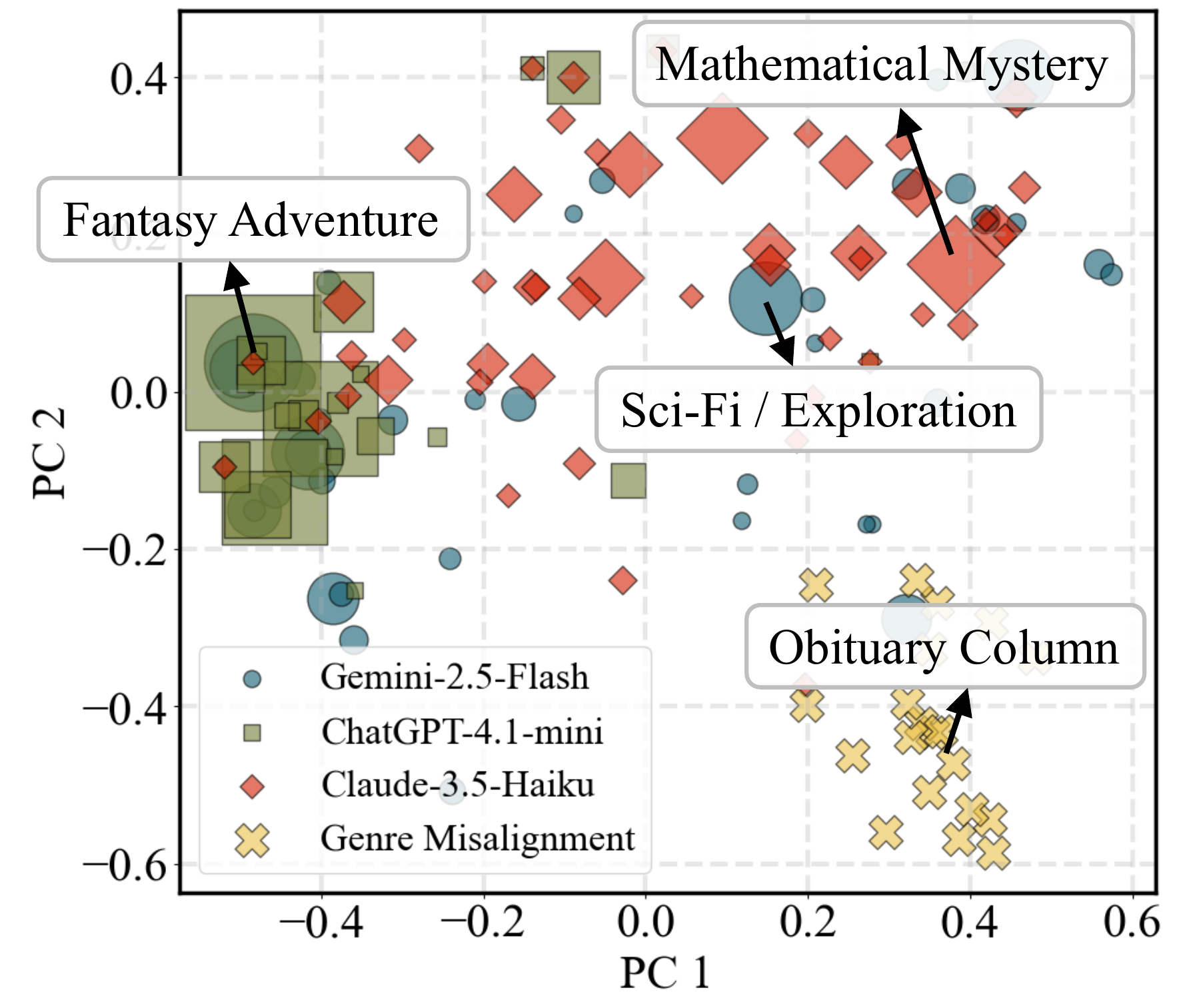}
        \caption{PCA visualization of genre embeddings.}
        \label{fig:pca}
    \end{subfigure}
    \caption{\textbf{Narrative genre preferences across models.} (a) The detailed genres with their ratio for each model; (b) PCA visualization of text embeddings of genre names selected by each model.}
    \label{fig:genre_misalignment}
\end{figure}

\subsection{Genre alignment drives performance}
\label{sec:5.4}

Information is understood differently depending on its style or framing~\citep{human:gentner1983structure, human:thibodeau2011metaphors}. We identify narrative genre as a primary factor that shapes the style and structure of problem descriptions. Figure~\ref{fig:genre_misalignment} shows the distribution of genres selected by the three closed-source models in Section~\ref{sec:experiments} across all benchmarks. Gemini-2.5-Flash and ChatGPT-4.1-mini choose genres such as ``Fantasy Adventure,'' whereas Claude-3.5-Haiku favors ``Mathematical Mystery.'' This suggests that each model develops its own preferred genre cluster within the narrative space.

To examine whether this genre preference reflects a functionally meaningful structure, we deliberately replace these well-aligned genres with incongruent ones. To this end, we manually curated $\mathcal{G}_{\text{mis}}$, a set of 20 misaligned genres with administrative, legal, or memorial characteristics, disjoint from the genres naturally favored by the models. While $f_{\text{narr}}$ selects genres freely based on $Q_i$ in the standard setting, we instead enforce the genre $g_{\text{mis}}$ to be drawn from $\mathcal{G}_{\text{mis}}$, obtaining \textbf{Misaligned Narratives} $\mathcal{N}_i^{j, \text{mis}} \sim P(\,\cdot \mid a_i^j, g_{\text{mis}})$. Details are provided in Appendix~\ref{appendix:misaligned_genres}.

As shown in Figure~\ref{fig:permuted_and_misaligned}, misaligned narratives show reduced performance compared to complete narratives across both benchmarks. This indicates that not all narratives are equally effective and that LLMs perform best when prompts align with genres conducive to reasoning. These findings suggest that LLMs recognize an optimal narrative space, and that \emph{representation alignment between problem description and model reasoning is a key factor in effective code generation.}

\begin{table}[t]
\centering
\small
\caption{\textbf{AST-based structural properties} of correct solutions generated by repeated sampling~(RS) and narrative prompting~(Narr.). ($^{**}p<0.01$, $^{***}p<0.001$)}
\setlength{\tabcolsep}{.5em}
\begin{tabular}{l|cc|cc}
\toprule
 & \multicolumn{2}{c|}{LiveCodeBench} & \multicolumn{2}{c}{CodeForces} \\
Metric & RS & Narr. & RS & Narr. \\
\midrule
Avg.\ functions    & 0.901 & 0.978$^{**}$  & 1.820 & 2.338$^{***}$ \\
Helper func.\ rate & 39.7\% & 39.8\%       & 41.4\% & 48.6\%$^{**}$ \\
AST depth          & 9.061 & 9.293$^{**}$  & 11.033 & 11.272$^{***}$ \\
\bottomrule
\end{tabular}
\label{tab:ast_analysis}
\end{table}

\subsection{Code-level structural analysis}
\label{sec:5.5}

While pass@$k$ and error decomposition measure functional correctness and algorithmic alignment, they do not directly measure whether narrative prompting induces more structured code at the implementation level. To address this, we extract the abstract syntax tree (AST) of each correct solution and compute three structural metrics.

We exclude HumanEval from this analysis, as its format requires completing code within a single predefined function, which artificially constrains structural variation. For LiveCodeBench and CodeForces, we compute: (i) Average functions, the mean number of function definitions per solution; (ii) Helper function rate, the proportion of solutions containing at least two functions or a nested function definition; and (iii) AST depth, the maximum depth of the tree. The statistical significance is assessed using a one-sided Mann--Whitney $U$ test.

As shown in Table~\ref{tab:ast_analysis}, our method induces more frequent decomposition into sub-functions, higher helper function usage, and deeper structural hierarchies across both benchmarks. These results complement the algorithm agreement analysis in Section~\ref{sec:5.2}: while that analysis shows narrative prompting guides models toward the correct algorithmic strategy, the AST-level results further demonstrate that \emph{it also encourages more modular and stepwise implementation}, suggesting that narrative reformulation improves code generation at both the algorithmic and structural levels.
\vspace{1em}
\section{Conclusion}
In this work, we proposed \textsc{StoryCoder}, a framework that reformulates coding problems into coherent narratives to promote integrative reasoning in LLMs, showing consistent performance gains across diverse benchmarks. Beyond demonstrating improved coverage and algorithmic alignment, our findings suggest that narrative coherence and representation alignment are key factors that shape problem-solving effectiveness. More broadly, our study shows the role of narratives as guiding frameworks that help organize and contextualize complex tasks. We expect that future work will explore adaptive genre selection, automated narrative refinement, and the extension of narrative-based prompting to domains such as mathematics, multimodal reasoning, and scientific discovery.
\section{Limitations}

\noindent\textbf{Method limitations.} The effectiveness of \textsc{StoryCoder} depends in part on the expressive and instruction-following capacity of the narrative generator, particularly for open-source models. Performance gains are also smaller on simpler benchmarks such as HumanEval, where high-level narrative abstraction provides less benefit. Overall, the method is most effective when such reformulation is both feasible and meaningful.

\noindent\textbf{Scope of applicability.} The proposed method is designed for competitive programming tasks with well-defined inputs, outputs, and constraints, and has not yet been validated on open-ended software engineering tasks or larger repository-level settings. Furthermore, defining what constitutes an optimal narrative transformation remains an open question. Future work could explore principled criteria for evaluating narrative quality beyond functional correctness.

\noindent\textbf{Data and evaluation.} This study evaluates performance primarily through execution-based benchmarks using the pass@$k$ metric. While we complement this with an AST-based analysis to assess structural patterns, evaluation using human judgment or software quality metrics remains for future work.

\section*{Acknowledgment}
This work was supported by the Institute of Information $\&$ Communications Technology Planning $\&$ Evaluation (IITP) grant funded by the Korea government (MSIT) [RS-2021-II211341, Artificial Intelligence Graduate School Program (Chung-Ang University) and RS-2022-II220124, Development of Artificial Intelligence Technology for Self-Improving Competency-Aware Learning Capabilities]. SNUAILAB, corp, supports this work.
\bibliography{custom}
\setcounter{table}{0}
\renewcommand{\thetable}{\Alph{table}}
\setcounter{figure}{0}
\renewcommand{\thefigure}{\Alph{figure}}
\setcounter{section}{0}
\renewcommand\thesection{\Alph{section}}
\appendix

\begin{table*}[t]
\caption{\textbf{Pass@10 performance} of open-source solvers ($f_{\text{solve}}$) using repeated sampling (RS), narratives generated by GPT-4.1-mini (G), and Claude-3.5-Haiku (C) as $f_{\text{narr}}$. While the degree of improvement varies across models, performance consistently improves even when $f_{\text{narr}}$ is a closed-source model other than Gemini-2.5-Flash, demonstrating the generalization of narrative reformulation across closed- and open-source models.}
\centering
\small
\begin{tabular}{l|ccc|ccc|ccc}
\toprule
 & \multicolumn{3}{c|}{HumanEval} & \multicolumn{3}{c|}{LiveCodeBench} & \multicolumn{3}{c}{CodeForces} \\
Model & RS & G & C & RS & G & C & RS & G & C  \\
\midrule
DSCoder 6.7B       & 82.86 & 85.71 & \textbf{86.67} & 22.29 & \textbf{25.71} & 22.29 & 13.12 & \textbf{13.40} & 11.38 \\
DSCoder V2 Lite    & 78.10 & 87.62 & \textbf{90.48} & 28.57 & 28.57 & \textbf{32.00} & 25.16 & 26.03 & \textbf{30.05} \\
Llama 3.1 8B       & 79.05 & \textbf{81.90} & 76.19 & 21.14 & \textbf{25.14} & 24.00 &  8.11 & 13.70 & \textbf{14.33} \\
Gemma 2 9B         & 63.81 & 67.62 & \textbf{83.81} & 20.00 & \textbf{22.86} & \textbf{22.86} & 11.69 & 16.06 & \textbf{18.93} \\
Gemma 2 27B        & 76.19 & 80.00 & \textbf{87.62} & 27.43 & 28.00 & \textbf{29.14} & 22.38 & 27.18 & \textbf{27.24} \\
Qwen 2.5 Coder 7B  & 89.52 & 88.57 & \textbf{94.29} & 26.86 & \textbf{29.71} & 29.14 & 19.31 & 20.61 & \textbf{22.35} \\
Qwen 2.5 Coder 32B & 92.38 & 88.57 & \textbf{94.29} & 30.86 & 34.29 & \textbf{37.14} & 15.08 & 20.61 & \textbf{22.81} \\
Mistral Small 24B  & 88.57 & 87.62 & \textbf{89.52} & 33.71 & \textbf{34.86} & 34.29 & 29.23 & \textbf{36.09} & 33.75 \\
\rowcolor{gray!15}
Average            & 81.31 & 83.45 & \textbf{87.86} & 26.36 & 28.64 & \textbf{28.86} & 18.01 & 21.71 & \textbf{22.61} \\
\bottomrule
\end{tabular}
\label{tab:gpt_and_claude_open_source}
\end{table*}

\newpage
\noindent{\Large \textbf{Appendix}} 
\label{appendix}

\section{Additional results}
\label{appendix:additional_results}

\subsection{Comprehensive results}
\label{appendix:comprehensive_results}

\textbf{Generalization to other narrative generators.} Table~\ref{tab:gpt_and_claude_open_source} reports additional results when GPT-4.1-mini or Claude-3.5-Haiku are used as $f_{\text{narr}}$ instead of Gemini-2.5-Flash, while the solver models are open-source. Similar to the main results in Table~\ref{tab:main_scores}, we observe consistent improvements over the Repeated Sampling (RS) prompts across open-source solvers, showing that the generalization of narrative reformulation is not limited to specific model but extends to other closed-source generators as well.

\begin{table*}[t]
\caption{\textbf{Pass@5 performance} of the narrative-only (Narr.) and narrative-original concatenation (Orig. + Narr.) settings. The results show consistent improvements over the baseline. For the first three closed-source models, we use the self-solving setting where $f_{\text{narr}} = f_{\text{solve}}$. For the following eight open-source models, $f_{\text{narr}}$ is fixed to Gemini-2.5-Flash.}
\centering
\small
\begin{tabular}{l|ccc|ccc|ccc}
\toprule
 & \multicolumn{3}{c|}{HumanEval} & \multicolumn{3}{c|}{LiveCodeBench} & \multicolumn{3}{c}{CodeForces} \\
Model & RS & \makecell{Narr.\\Only} & \makecell{Orig.\\+ Narr.} & RS & \makecell{Narr.\\Only} & \makecell{Orig.\\+ Narr.} & RS & \makecell{Narr.\\Only} & \makecell{Orig.\\+ Narr.}  \\
\midrule
Gemini-2.5-Flash & 95.93 & \textbf{96.19} & \textbf{96.19} & 50.16 & 50.86 & \textbf{52.57} & 52.24 & \textbf{58.87} & 53.59 \\
GPT-4.1-mini     & \textbf{95.23} & 94.29 & 94.29 & 48.22 & \textbf{53.71} & 49.14 & 34.64 & \textbf{42.27} & 35.47 \\
Claude-3.5-Haiku & 84.97 & 87.62 & \textbf{93.33} & 31.97 & 28.57 & \textbf{34.86} & 44.61 & 32.45 & \textbf{48.68} \\
\rowcolor{gray!15}
Average            & 92.04 & 92.70 & \textbf{94.60} & 43.45 & 44.38 & \textbf{45.52} & 43.83 & 44.53 & \textbf{45.91} \\
\midrule
DSCoder 6.7B       & 80.95 & 81.90 & \textbf{88.57} & 21.14 & 21.71 & \textbf{25.71} & 11.23 & 12.30 & \textbf{14.24} \\
DSCoder V2 Lite    & 78.10 & 87.62 & \textbf{92.38} & 27.43 & 29.14 & \textbf{33.14} & 21.56 & 24.69 & \textbf{31.12} \\
Llama 3.1 8B       & \textbf{79.05} & 57.14 & 75.24 & 20.00 & \textbf{26.29} & 25.71 & 7.16 & \textbf{14.82} & 12.30 \\
Gemma 2 9B         & 63.81 & 78.10 & \textbf{80.00} & 20.00 & 22.29 & \textbf{24.00} & 10.42 & 16.06 & \textbf{18.58} \\
Gemma 2 27B        & 76.19 & \textbf{86.67} & \textbf{86.67} & 26.29 & 29.71 & \textbf{32.57} & 20.96 & 24.40 & \textbf{26.42} \\
Qwen 2.5 Coder 7B  & 88.57 & \textbf{92.38} & \textbf{92.38} & 25.71 & 28.57 & \textbf{30.86} & 16.96 & 20.46 & \textbf{23.28} \\
Qwen 2.5 Coder 32B & 92.38 & \textbf{94.29} & 93.33 & 29.71 & 34.86 & \textbf{35.43} & 11.46 & \textbf{24.08} & 14.82 \\
Mistral Small 24B  & 86.67 & 90.48 & \textbf{93.33} & 32.00 & \textbf{32.57} & \textbf{32.57} & 24.86 & \textbf{34.12} & 33.14 \\
\rowcolor{gray!15}
Average            & 80.72 & 83.57 & \textbf{87.74} & 25.29 & 28.14 & \textbf{30.00} & 15.58 & 21.37 & \textbf{21.74} \\
\bottomrule
\end{tabular}
\label{tab:pass5_closed_source}
\end{table*}

\begin{table}[t]
    \centering
    \small
    \caption{
        \textbf{Pass@5 performance} on LiveCodeBench-v6 for latest models. Narrative prompting consistently outperforms repeated sampling (RS) across all models.
    }
    \begin{tabular}{l|cc}
    \toprule
    Model & RS & \makecell{Orig.\\+ Narr.} \\
    \midrule
    Gemini-3.1-Flash-Lite-Preview & 61.14 & \textbf{61.71} \\
    GPT-5.4-mini                  & 51.43 & \textbf{53.71} \\
    Claude-Sonnet-4.6             & 65.14 & \textbf{66.86} \\
    \bottomrule
    \end{tabular}
    \label{tab:recent_models}
\end{table}

\noindent\textbf{Comparison of narrative-only and concatenated variants.} We report pass@10 in the main paper, which aggregates performance over ten responses per problem (five variants of narrative-only and five variants of narrative concatenated with the original question). Here, we report the pass@5 results for each of the five responses in each setting individually, as shown in Table~\ref{tab:pass5_closed_source}. Consequently, the RS column also reports the pass@5 score computed from five responses. As shown in the table, both the variants of narrative-only and concatenation settings consistently outperform the baseline. Surprisingly, we observe cases across multiple models and benchmarks where narrative-only inputs outperform the concatenated form. This suggests that certain elements of the original problem statement may act as distractors that hinder correct reasoning or encourage spurious shortcuts, and it emphasizes the importance of \mbox{\textsc{StoryCoder}} in reinforcing step-by-step reasoning.

\noindent\textbf{Generalization to recent models.} Table~\ref{tab:recent_models} reports pass@5 results on LiveCodeBench for three recently released models: Gemini-3.1-Flash-Lite-Preview, GPT-5.4-mini, and Claude-Sonnet-4-6. Narrative prompting consistently outperforms repeated sampling across all three models, demonstrating that the effectiveness of \mbox{\textsc{StoryCoder}} generalizes to the latest models available at the time of writing, beyond those evaluated in the main experiments.

\subsection{Additional ablation study}
\label{appendix:ablation_study}

In the main pipeline, algorithm and genre tags serve as additional organizational cues for narrative generation, and these labels operate solely within the generator $f_{\text{narr}}$, which means they play only an indirect role in the overall problem-solving process. To isolate the effect of the structural transformation itself, we also evaluate a variant that omits these tags, which we refer to as No-Tag Narrative.

The No-Tag Narrative is produced by applying the same transformation guidelines while simply removing the algorithm and genre sections. As shown in Table~\ref{tab:algorithm_and_genre_free_results}, this variant still improves over the baseline and performs particularly well on easier benchmarks such as HumanEval. In contrast, the full narrative tends to perform better on more challenging benchmarks, suggesting that $f_{\text{narr}}$ can benefit from exploring the algorithmic and genre space when composing narratives for more complex problems. This pattern indicates that the primary benefit comes from structural reorganization, while the auxiliary tags help $f_{\text{narr}}$ better guide difficult problem spaces.

\subsection{Comparison with other baselines}
\label{appendix:comparison_with_other_baselines}

To examine how different degrees of problem reformulation affect code generation, we compare \mbox{\textsc{StoryCoder}} with paraphrasing, its concatenated variant (PC), and Story-of-Thought (SoT)~\citep{sot} in Table~\ref{tab:paraphrase_and_its_concat}. Paraphrasing modifies only surface expressions without altering the underlying structure, and PC further increases input length by concatenating five paraphrases without introducing new algorithmic cues. SoT applies open-ended narrative prompting but relies on guidelines designed for knowledge-intensive multiple-choice tasks rather than algorithmic problem solving. Consequently, none of these methods yields consistent improvements across benchmarks, whereas \mbox{\textsc{StoryCoder}} consistently achieves the strongest performance across all benchmarks. Figure~\ref{fig:appendix_transformation_example_paraphrase} shows that expression-level rewriting alone do not fundamentally enhance reasoning.

\begin{table*}[t]
\caption{\textbf{Pass@10 Performance without algorithm or genre tags.} Narr. (No-Tag) denotes the results obtained using a narrative generated without algorithm or genre tags, while Narr. denotes the tagged version used in the main pipeline.}
\centering
\small
\setlength{\tabcolsep}{5pt}
\begin{tabular}{l|ccc|ccc|ccc}
\toprule
 & \multicolumn{3}{c|}{HumanEval} & \multicolumn{3}{c|}{LiveCodeBench} & \multicolumn{3}{c}{CodeForces} \\
Model & RS & \makecell{Narr.\\(No-Tag)} & Narr. & RS & \makecell{Narr.\\(No-Tag)} & Narr. & RS & \makecell{Narr.\\(No-Tag)} & Narr. \\
\midrule
Gemini-2.5-Flash   & 96.19 & \textbf{97.14} & 96.19 & 49.71 & \textbf{58.29} & 57.14 & 60.00 & 64.09 & \textbf{67.55} \\
\midrule
DSCoder 6.7B       & 82.86 & \textbf{91.43} & 90.48 & 22.29 & 24.57 & \textbf{27.43} & 13.12 & 14.36 & \textbf{18.00} \\
DSCoder V2 Lite    & 78.10 & \textbf{93.33} & \textbf{93.33} & 28.57 & 33.71 & \textbf{34.29} & 25.16 & \textbf{36.44} & 33.14 \\
Llama 3.1 8B       & 79.05 & \textbf{88.57} & 81.90 & 21.14 & 24.57 & \textbf{27.43} & 8.11 & 17.31 & \textbf{19.08} \\
Gemma 2 9B         & 63.81 & \textbf{87.62} & 82.86 & 20.00 & 25.14 & \textbf{26.29} & 11.69 & 19.25 & \textbf{21.39} \\
Gemma 2 27B        & 76.19 & \textbf{89.52} & 87.62 & 27.43 & 30.86 & \textbf{34.29} & 22.38 & 29.73 & \textbf{30.97} \\
Qwen 2.5 Coder 7B  & 89.52 & \textbf{93.33} & \textbf{93.33} & 26.86 & 31.43 & \textbf{33.14} & 19.30 & \textbf{27.67} & 26.74 \\
Qwen 2.5 Coder 32B & 92.38 & \textbf{95.24} & 94.29 & 30.86 & 38.86 & \textbf{40.00} & 15.08 & \textbf{29.26} & 27.10 \\
Mistral Small 24B  & 88.57 & \textbf{96.19} & 94.29 & 33.71 & \textbf{36.57} & 34.86 & 29.23 & 41.45 & \textbf{42.86} \\
\rowcolor{gray!15}
Average            & 81.31 & \textbf{91.90} & 89.76 & 26.36 & 30.71 & \textbf{32.22} & 18.01 & 26.93 & \textbf{27.41} \\
\bottomrule
\end{tabular}
\label{tab:algorithm_and_genre_free_results}
\end{table*}

\subsection{\texorpdfstring{\boldmath Pass@$k$}{Pass@k} curves on three benchmarks}
\label{appendix:pass_k_curve}

\begin{figure*}[t]
  \centering
  \begin{subfigure}{1.0\textwidth}
    \centering
    \includegraphics[width=\linewidth]{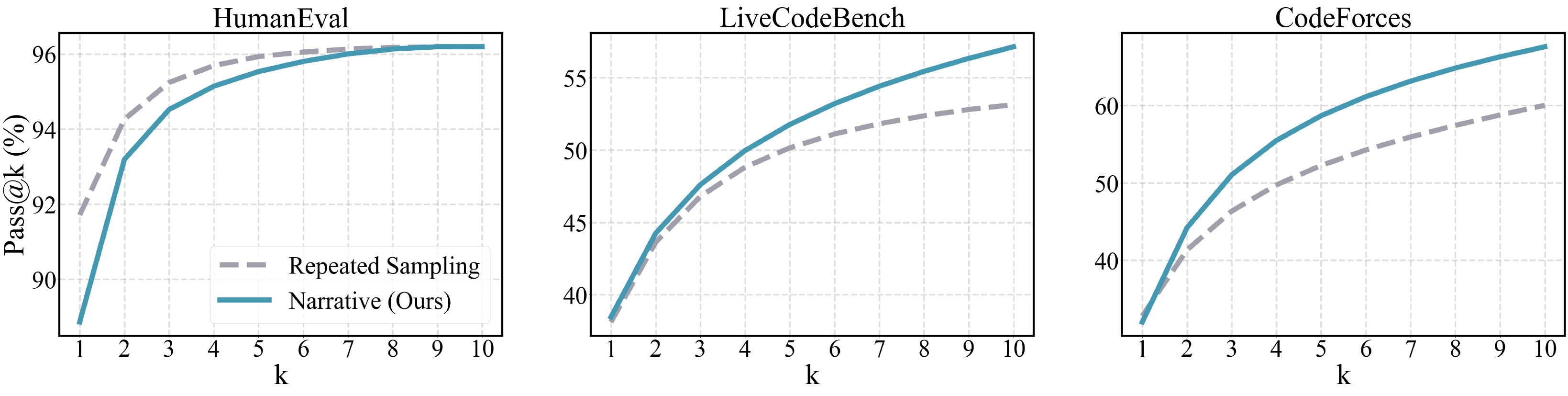}
    \caption{Gemini-2.5-Flash}
  \end{subfigure}

  \vspace{0.1cm}

  \begin{subfigure}{1.0\textwidth}
    \centering
    \includegraphics[width=\linewidth]{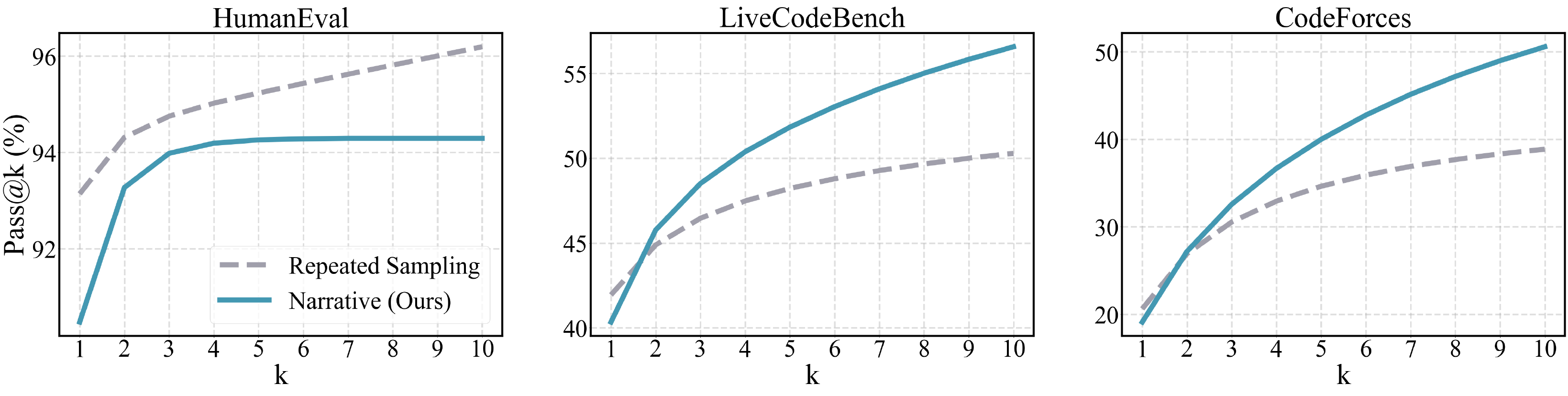}
    \caption{ChatGPT-4.1-mini}
  \end{subfigure}
  
  \vspace{0.1cm}

  \begin{subfigure}{1.0\textwidth}
    \centering
    \includegraphics[width=\linewidth]{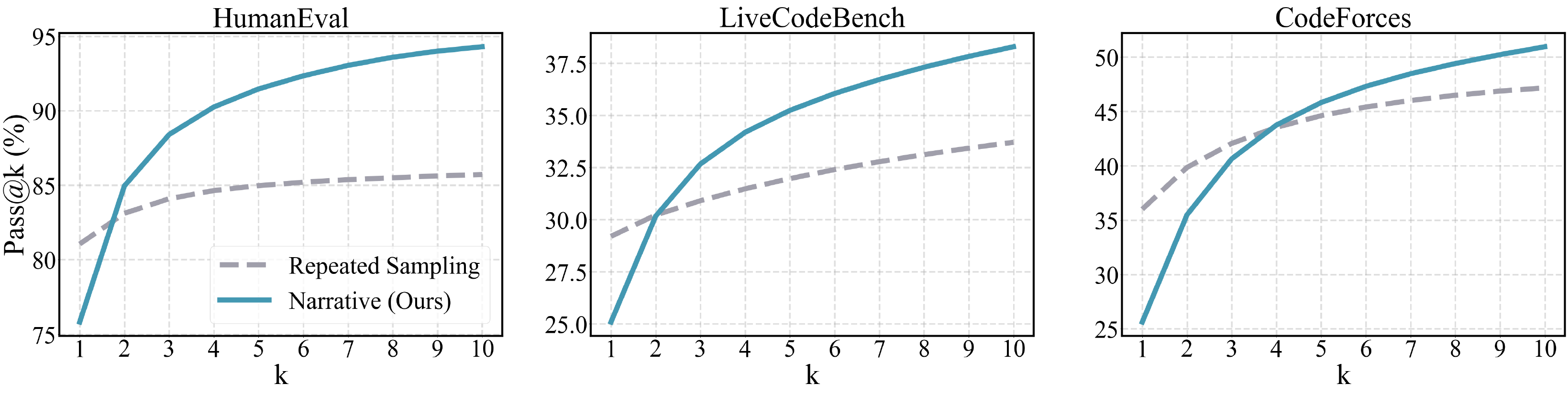}
    \caption{Claude-3.5-Haiku}
  \end{subfigure}
  
  \caption{\textbf{Pass@k performance} of closed-source models for $k=1, \ldots, 10$. Except for ChatGPT-4.1-mini on HumanEval, narrative prompting consistently outperforms the baseline as $k$ increases across all models and benchmarks.}
  \label{fig:appendix_pass_k_closed_models}
\end{figure*}

\begin{figure*}[ht]
    \centering
    \includegraphics[width=0.92\textwidth]{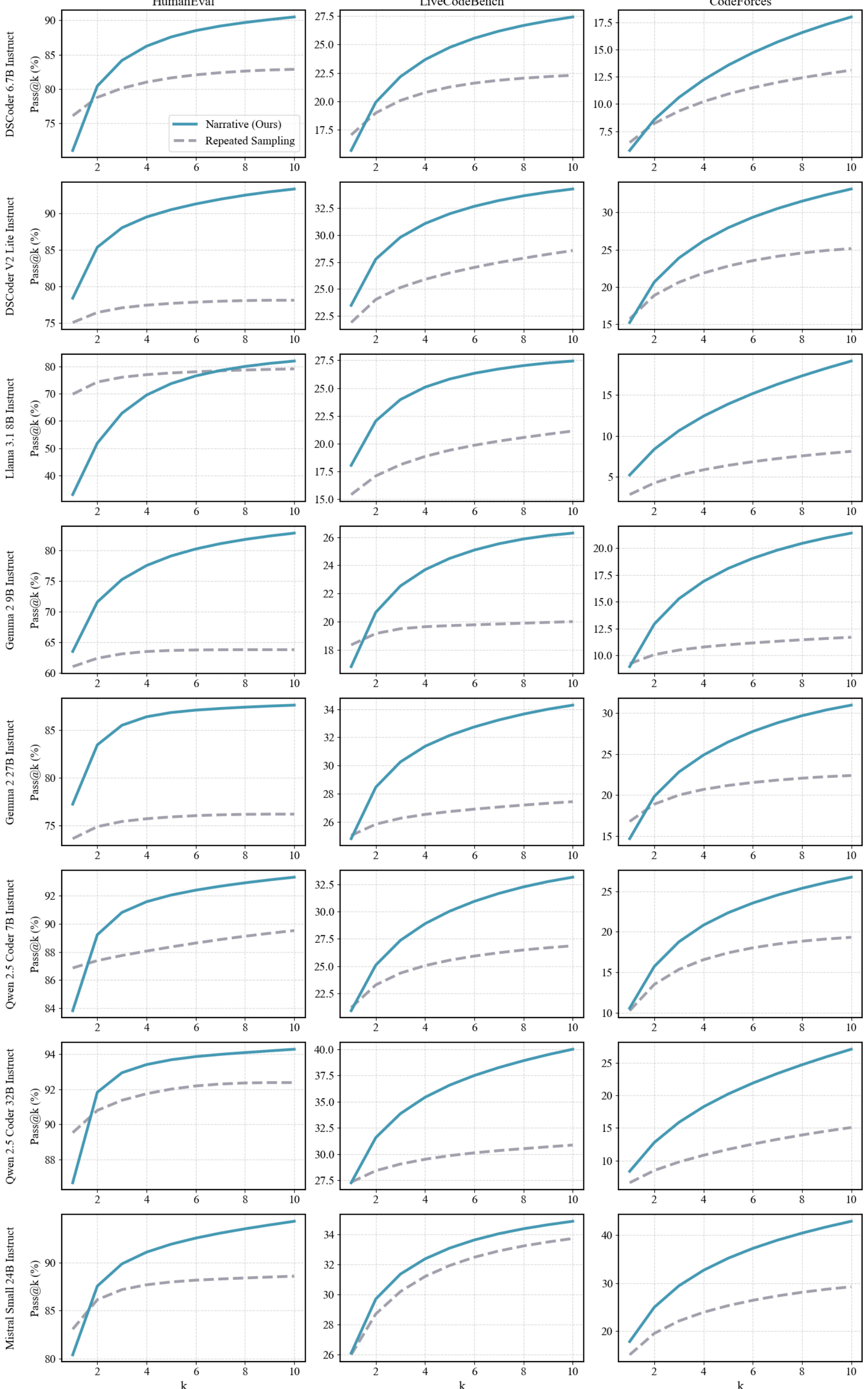}
    \caption{\textbf{Pass@k performance} of open-source models for $k=1, \ldots, 10$. Across all models and benchmarks, narrative prompting consistently outperforms the baseline as $k$ increases.}
    \label{fig:appendix_all_open_source}
\end{figure*}

Figures~\ref{fig:appendix_pass_k_closed_models} and \ref{fig:appendix_all_open_source} present the pass@k curves on HumanEval~\citep{humaneval}, LiveCodeBench~\citep{livecodebench}, and CodeForces~\citep{codeforces} for closed-source and open-source models. Except for small values of $k$ (around $k=1$ to $4$), narrative prompting outperforms the baseline (Repeated Sampling) in all cases. As $k$ increases, the performance gains become smaller on the easier HumanEval benchmark, while they continue to grow on the more challenging LiveCodeBench and CodeForces benchmarks.

\section{Experimental details}

\subsection{Evaluation metric}
\label{appendix:evaluation_metric}
The pass@$k$ metric evaluates the probability that at least one correct solution is obtained among $k$ independently sampled outputs. Formally, given $n$ generated outputs with $c$ correct ones, the expected success rate is

\begin{equation}
\text{pass@}k = \mathbb{E}\left[ 1 - \frac{\binom{n-c}{k}}{\binom{n}{k}} \right].
\label{eq:appendix_passk}
\end{equation}

In code generation benchmarks, there is no single canonical solution, and multiple programs may be valid for the same problem. As a result, evaluation is typically performed by repeated sampling and checking whether at least one passes all test cases. This characteristic naturally motivates the use of pass@$k$, which measures the probability that a model produces at least one correct solution within $k$ attempts. Consequently, pass@$k$ has become an intuitive and widely adopted metric in practice.

\renewcommand{\thefootnote}{\arabic{footnote}}

\subsection{Dataset filtering}
\label{appendix:dataset_filtering}

For reproducibility, we summarize the dataset filtering process applied to each benchmark.

\noindent\textbf{HumanEval}~\citep{humaneval}\textbf{.}\footnote{\url{https://github.com/openai/human-eval}} We exclude samples without input/output examples and standardize the format of all examples. Samples without reliably identifiable input/output examples (e.g., missing a function-name usage example) cannot be evaluated under our execution-based framework and are therefore excluded. Function names in signatures and examples are unified when they differ. After these adjustments, we obtain a filtered set of 105 samples.

\noindent\textbf{LiveCodeBench}~\citep{livecodebench}\textbf{.}\footnote{\url{https://github.com/LiveCodeBench/LiveCodeBench}} To avoid data contamination, we use the release-v6 subset that covers samples from January to April 2025. This subset consists of 112 samples from AtCoder~\citep{atcoder} and 63 from LeetCode~\citep{leetcode}, amounting to 175 samples.

\noindent\textbf{CodeForces}~\citep{codeforces}\textbf{.}\footnote{\url{https://huggingface.co/datasets/open-r1/codeforces}} To keep the experiment computationally feasible given that the full dataset contains more than 10,000 problems, we select the tasks with moderate text length (length $\leq 1000$). We further exclude problems without input/output examples and retain only intermediate- and advanced-level tasks (rating $\geq 2000$). This filtering process yields a final set of 265 samples.

To verify that \mbox{\textsc{StoryCoder}} is robust to long problem statements, we additionally conduct experiments on CodeForces problems with substantially longer descriptions. Keeping all other settings unchanged, we extract all samples with text length greater than 1,000 and randomly select 128 of them to construct a subset, CodeForces-L. As shown in Table~\ref{tab:codeforces_l}, \mbox{\textsc{StoryCoder}} consistently outperforms the baseline even on this long-text subset, demonstrating its robustness to the length of problem descriptions.

\begin{table*}[t]
\caption{\textbf{Pass@10 Performance} comparison of Paraphrase (Para.), Paraphrase Concatenation (PC), and Story-of-Thought (SoT) prompts. Paraphrase alters the surface expressions while preserving its meaning. PC concatenates five paraphrase variants per question to intentionally increase prompt length. SoT applies open-ended narrative prompting but relies on guidelines designed for knowledge-intensive multiple-choice tasks rather than algorithmic problem solving. \mbox{\textsc{StoryCoder}} consistently outperforms all three methods, showing that its improvements comes from structured narrative reformulation grounded in algorithmic reasoning, rather than surface-level expression changes, increased input length, or task-agnostic narrative guidelines.}

\centering
\small
\tabcolsep=.5em
\begin{tabular}{l|cccc|cccc|cccc}
\toprule
 & \multicolumn{4}{c|}{HumanEval} & \multicolumn{4}{c|}{LiveCodeBench} & \multicolumn{4}{c}{CodeForces} \\
Model & Para. & PC & SoT & Narr. & Para. & PC & SoT & Narr. & Para. & PC & SoT & Narr. \\
\midrule
Gemini-2.5-Flash   & 94.29 & 93.07 & 94.29 & \textbf{96.19} & 50.29 & 50.29 & 50.86 & \textbf{57.14} & 50.51 & 51.55 & 61.86 & \textbf{67.55} \\
\midrule
DSCoder 6.7B       & 81.90 & 85.15 & 87.62 & \textbf{90.48} & 24.00 & 24.00 & 26.29 & \textbf{27.43} & 11.67 & 13.90 & 16.10 & \textbf{18.01} \\
DSCoder V2 Lite    & 85.71 & 88.12 & 90.48 & \textbf{93.33} & 28.57 & 30.29 & 28.00 & \textbf{34.29} & 26.40 & 27.67 & 31.40 & \textbf{33.14} \\
Llama 3.1 8B       & 83.81 & 86.14 & \textbf{86.67} & 81.90 & 22.86 & 24.57 & \textbf{28.57} & 27.43 & 7.62 & 13.70 & 18.26 & \textbf{19.08} \\
Gemma 2 9B         & 68.57 & 70.30 & 80.95 & \textbf{82.86} & 20.57 & 20.00 & 22.86 & \textbf{26.29} & 13.40 & 14.04 & 16.76 & \textbf{21.39} \\
Gemma 2 27B        & 81.90 & 79.21 & 84.76 & \textbf{87.62} & 27.43 & 26.29 & 28.00 & \textbf{34.29} & 22.49 & 21.42 & 26.92 & \textbf{30.97} \\
Qwen 2.5 Coder 7B  & 88.57 & 88.12 & 92.38 & \textbf{93.33} & 26.86 & 28.57 & 32.00 & \textbf{33.14} & 23.28 & 25.59 & 25.01 & \textbf{26.74} \\
Qwen 2.5 Coder 32B & 92.38 & 91.09 & 93.33 & \textbf{94.29} & 34.29 & 33.71 & 38.29 & \textbf{40.00} & 19.63 & 17.60 & \textbf{33.03} & 27.10 \\
Mistral Small 24B  & 87.62 & 90.10 & 92.38 & \textbf{94.29} & 33.14 & 33.71 & \textbf{34.86} & \textbf{34.86} & 29.53 & 34.24 & 40.78 & \textbf{42.87} \\
\rowcolor{gray!15}
Average            & 83.81 & 84.78 & 88.57 & \textbf{89.76} & 27.22 & 27.64 & 29.86 & \textbf{32.22} & 19.25 & 21.02 & 26.03 & \textbf{27.41} \\
\bottomrule
\end{tabular}
\label{tab:paraphrase_and_its_concat}
\end{table*}

\begin{table}[t]
\centering
\small
\caption{\textbf{Pass@10 performance} on CodeForces-L (longer descriptions).
The table reports results where $f_\text{narr}$ is fixed to Gemini-2.5-Flash
and $f_\text{solve}$ varies by row. Narrative prompting consistently outperforms the baseline across subsets with longer problem lengths.}
\vspace{-0.5em}
\begin{tabular}{l|cc}
\toprule
 & \multicolumn{2}{c}{CodeForces-L} \\
Model & RS & Narr. \\
\midrule
Gemini-2.5-Flash & 35.94 & \textbf{53.91} \\
\midrule
DSCoder 6.7B       & 1.56 & \textbf{7.03} \\
DSCoder V2 Lite    & 8.59 & \textbf{18.75} \\
Llama 3.1 8B       & 0.0 & \textbf{3.12} \\
Gemma 2 9B         & 2.34 & \textbf{7.81} \\
Gemma 2 27B        & 11.72 & \textbf{14.06} \\
Qwen 2.5 Coder 7B  & 7.81 & \textbf{10.94} \\
Qwen 2.5 Coder 32B & 9.38 & \textbf{16.41} \\
Mistral Small 24B  & 12.5 & \textbf{20.31} \\
\rowcolor{gray!15}
Average            & 6.74 & \textbf{12.30} \\
\bottomrule
\end{tabular}
\label{tab:codeforces_l}
\end{table}

\subsection{Narrative transformation examples}
\label{appendix:narrative_transformation_examples}

\begin{figure*}[ht]
    \centering
    \includegraphics[width=0.95\textwidth]{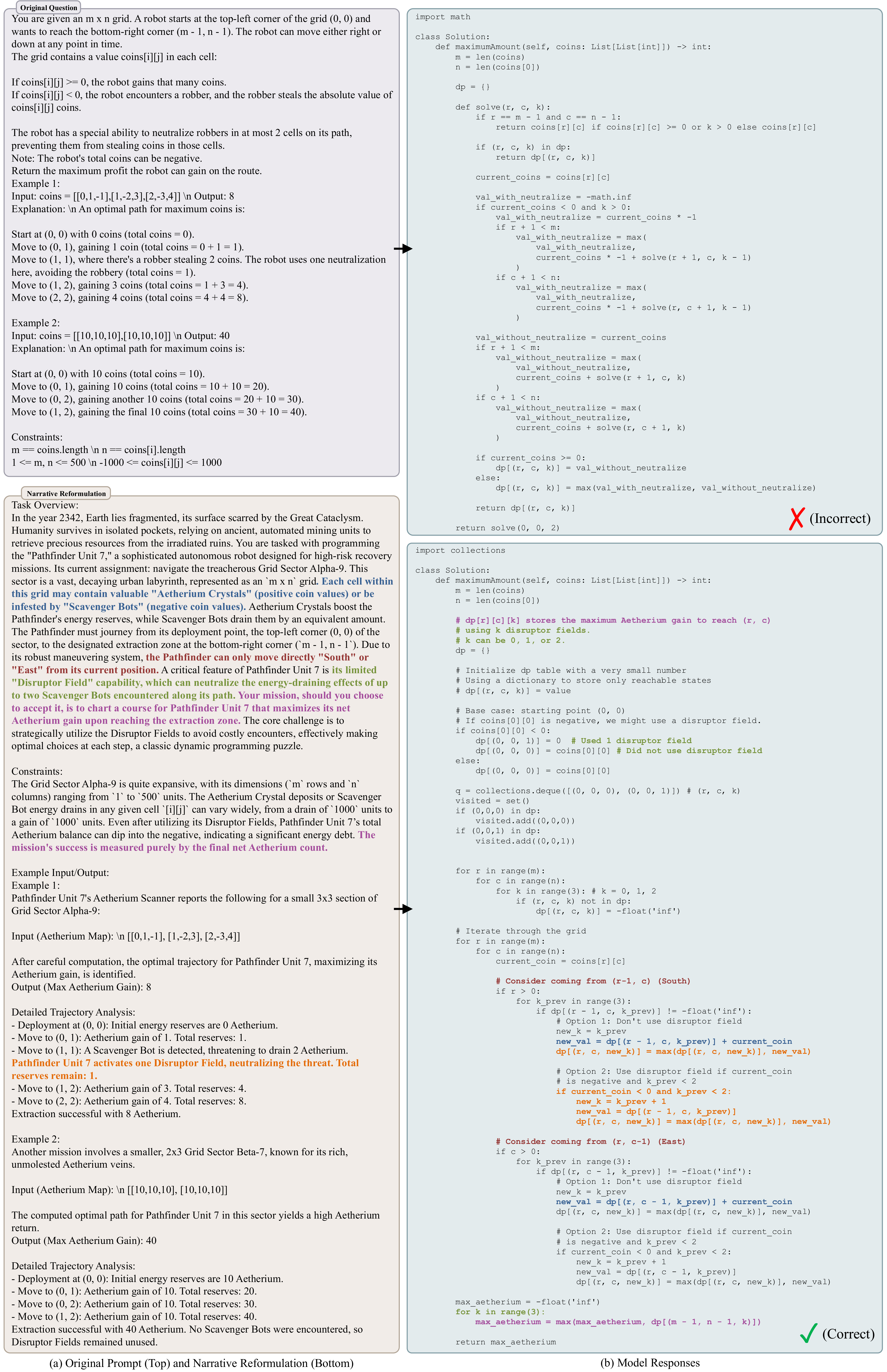}
    \caption{A complete example of a narrative reformulation ($f_\text{narr} = f_\text{solve} =$ Gemini-2.5-Flash), where $f_\text{solve}$ correctly implements the intended dynamic programming specification.}
    \label{fig:appendix_transformation_example}
\end{figure*}

\begin{figure*}[ht]
    \centering
    \includegraphics[width=1.0\textwidth]{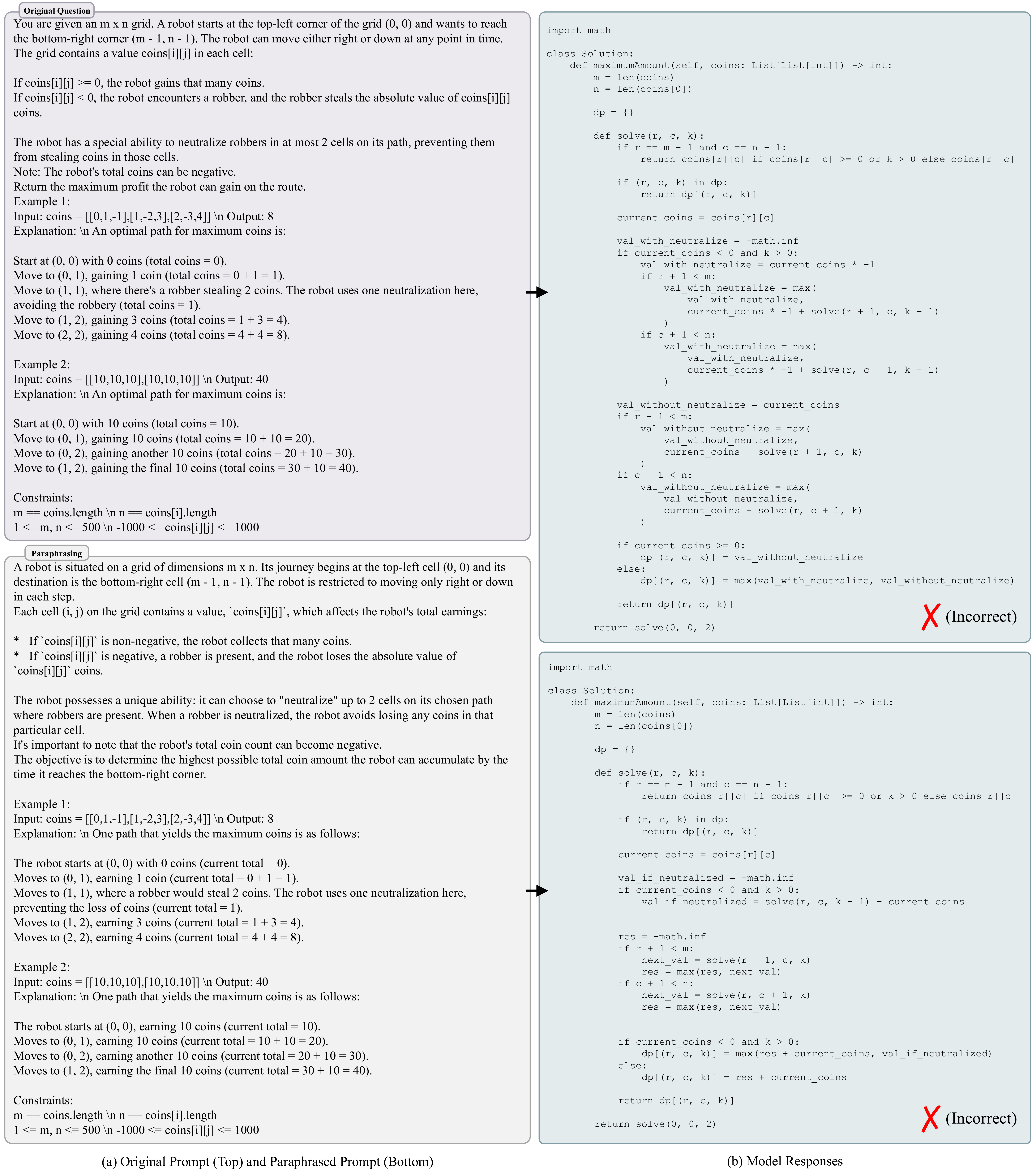}
    \caption{Paraphrasing example. Both the paraphrase and the solution were generated by Gemini-2.5-Flash. Unlike the narrative transformation in Figure~\ref{fig:appendix_transformation_example}, paraphrasing modifies only the surface wording while preserving the original structure, so it does not lead the model to think in a different way, which is what \mbox{\textsc{StoryCoder}} achieves.}
    \label{fig:appendix_transformation_example_paraphrase}
\end{figure*}

\textbf{How narrative reformulation guides reasoning.} Figure~\ref{fig:appendix_narrative_transformation_guidelines} shows the prompts for converting original coding questions into narrative format. Figure~\ref{fig:appendix_transformation_example} provides a complete example from LiveCodeBench, including the original coding problem, its narrative reformulation, and the responses generated by Gemini-2.5-Flash. This example illustrates how the narrative formulation helps the model solve the problem. In this example, \mbox{\textsc{StoryCoder}} makes the problem's core structure (path constraints, state branching, and the global optimization objective) explicit through narrative, guiding the LLM to construct the correct DP state space and transition rules. The narrative elements and their corresponding code segments are annotated in matching colors, and these structural cues align the model's reasoning to perform correct branching and global optimization, leading to the final correct solution.

\vspace{.5em}

\noindent\textbf{Intuitions behind the effectiveness of narratives.} Beyond these examples, we propose the following intuitions for why \mbox{\textsc{StoryCoder}} improves performance: (i) Narratives align better with the pretraining distribution of LLMs, which is dominated by descriptive and story-like text. Formally, let $\mathcal{L}(x)$ denote the average negative log-likelihood of the model on a sequence $x$. Since narrative text constitutes a substantially larger portion of pretraining corpora than code-formatted problem descriptions, we expect $\mathcal{L}(\mathcal{N}_i) < \mathcal{L}(Q_i)$, suggesting that narrative reformulation acts as a bridge connecting broad linguistic knowledge from pretraining to coding tasks. This is consistent with our empirical findings across 11 models; (ii) Narratives reorganize the problem into a clearer, more solvable structure by turning scattered constraints and abstract rules into a grounded, interpretable description that helps the model identify the appropriate algorithmic pattern; (iii) Narratives induce a more linear and model-friendly reasoning flow by outlining a natural step-by-step progression and reducing the model's tendency to take incorrect shortcuts during implementation.

\subsection{Narrative validity filtering}
\label{appendix:narrative_validity_filtering_criteria}

In Section~\ref{sec:4.2}, all three closed-source models generate valid narratives that satisfy the required format with 100\% accuracy. However, we observe that open-source models occasionally fail to produce valid narrative texts, which we attribute to limited instruction-following capabilities in smaller models. To ensure a fair and reliable evaluation, we apply the following filtering criteria: narrative outputs are considered invalid if (i) the sequence length is fewer than 50 tokens (near-empty content), or (ii) the sequence length exceeds 99\% of the model's maximum generation limit, which corresponds to degenerate token repetition, or (iii) the output lacks the required components (task overview, constraints, and input/output format). Illustrative examples of each invalid type are provided in Table~\ref{tab:open_source_invalid_narrative_examples}.

\begin{figure*}[t]
\centering
\begin{tcolorbox}[title=Narrative Transformation Guidelines]
\lstset{
  basicstyle=\ttfamily\small,
  breaklines=true,
  columns=fullflexible
}
\begin{lstlisting}
Please transform the coding problem into a narrative story using the following guidelines.

### Guidelines for Narrative Conversion:

Before writing the narrative, you must complete two preliminary steps:

1. Review the major categories of coding test algorithms:
   - Graph Algorithms
   - Dynamic Programming
   - Greedy Algorithms
   - Sorting and Searching
   - String Algorithms
   - Data Structures
   - Mathematics and Number Theory
   - Simulation and Implementation

2. Decide which algorithm category the given problem most closely belongs to.
   Then, select a narrative genre that naturally aligns with the chosen algorithm.

### Output Format:

You must write the output in the exact following order with the specified headers:

- Algorithm Category: (one of the categories above)

- Narrative Genre: (a fitting genre of your choice)

- Task Overview: Describe the background and objective of the problem in a clear, narrative-inspired manner. The selected algorithm should be introduced naturally here, with its logic explained as part of the setting or scenario.

- Constraints: State input sizes, value ranges, conditions, and key operational rules. If efficiency or time limits exist, express them as natural constraints. The chosen algorithm should also shape these rules.

- Example Input/Output: Reframe the examples as part of the scenario's flow. Present them as clear, contextual situations.

The narrative must include all essential parts of the original problem, ensuring no constraints, goals, or examples are omitted.
Do not include any other text outside these five sections.
Do not attempt to solve the problem or provide any code. Your task is only to transform the problem statement into the narrative format as specified.

### The coding problem is as follows:

{Coding Problem}
\end{lstlisting}
\end{tcolorbox}
\caption{Instructions for converting a code generation benchmark question into a narrative format.}
\label{fig:appendix_narrative_transformation_guidelines}
\end{figure*}

\subsection{Misaligned genres}
\label{appendix:misaligned_genres}

To construct the misalignment setting, we curated a set of genres that are intentionally incongruent with problem descriptions. These genres were selected to represent contexts that are stylistically or semantically distant from typical programming tasks, ensuring that the resulting narratives do not naturally align with the problem’s intent. Table~\ref{tab:misaligned_genres} presents the complete list of misaligned genres, grouped into four categories.

\begin{table*}[t]
\centering
\caption{Complete list of misaligned genres grouped into four categories.}
\label{tab:misaligned_genres}
\begin{tabular}{l l}
\toprule
\textbf{Category} & \textbf{Misaligned Genres} \\
\midrule
\makecell[{{p{2.6cm}}}]{
  Practical / \\ Administrative \\ Documents
} &
\makecell[{{p{10cm}}}]{
  Hospital Intake Form; Medical Prescription Form; \\ Personal Information Consent Form; Insurance Claim Form; \\ Visa Application Form; Tax Return Form
} \\
\midrule
\makecell[{{p{2.6cm}}}]{
  Legal / \\ Public Records
} &
\makecell[{{p{10cm}}}]{
  Court Transcript of an Extortion Case; Heavy Machinery Operator \\ License; Military Service Exemption Certificate; Divorce Decree; \\ Bank Loan Agreement
} \\
\midrule
\makecell[{{p{2.6cm}}}]{
  Industrial / \\ Media Contexts
} &
\makecell[{{p{10cm}}}]{
  Billboard Advertisement for a Toothbrush; Radio Weather Forecast; Model Agency Contract
} \\
\midrule
\makecell[{{p{2.6cm}}}]{
  Funerary / \\ Ritual Records
} &
\makecell[{{p{10cm}}}]{
  Funeral Service Program; Memorial Tribute Writing; Obituary Column; Eulogy; Gravestone Inscription; Condolence Letter
} \\
\bottomrule
\end{tabular}
\end{table*}

\begin{table*}[t]
\centering
\caption{Examples of invalid narratives generated by Llama 3.1 8B Instruct, Gemma 2 9B Instruct, and DeepSeek V2 Lite Chat, in table order. Invalid narrative types: (i) Incomplete Output: too short to construct a narrative; (ii) Repetitive Output: filled with meaningless repetition; (iii) Missing Components: lacking the three required components (task overview, constraints, and input/output format).}
\label{tab:open_source_invalid_narrative_examples}

\begin{tabular}{l l}
\toprule
\textbf{Invalid Type} & \textbf{Generated Text} \\
\midrule
\makecell[{{p{2cm}}}]{
  Incomplete \\ Output
} &
\makecell[{{p{13cm}}}]{
  ``return [x for x in l if x > 0]\textbackslash n\textbackslash n\textbackslash n\textbackslash n**Please note:** This is a formatted list, but the narrative should not mention \textbackslash "formatted list\textbackslash " or any specific format. \textbackslash n\textbackslash n- Algorithm Category: \textbackslash n\textbackslash n- Narrative Genre:\textbackslash n\textbackslash n- Task Overview:\textbackslash n\textbackslash n- Constraints:\textbackslash n\textbackslash n- Example Input/Output:''
} \\
\midrule
\makecell[{{p{2cm}}}]{
  Repetitive \\ Output
} &
\makecell[{{p{13cm}}}]{
  ``Here's the conversion to a narrative format:\textbackslash n\textbackslash n- Algorithm Category: Dynamic Programming\textbackslash n\textbackslash n- Narrative Genre: Mysterious Archaeology\textbackslash n\textbackslash n- Task Overview: In the heart of the mystical Amazon, an ancient temple holds secrets of the past. Your mission as a brilliant archaeologist is to uncover the temple's treasure, guarded by the mysterious priestesses. \textit{(omitted)} Conversely, when your opponent has a set of pieces including a 5, 3, -5, 2, -3, 3, 9, 0, 123, 1, -10, the king's value is 123.\textbackslash n\textbackslash n\textbackslash n\textbackslash n---\textbackslash n\textbackslash nHere is the transformed narrative in the specified format.\textbackslash n\textbackslash n\textbackslash n\textbackslash n---\textbackslash n\textbackslash nHere is the transformed narrative in the format you requested.\textbackslash n\textbackslash n\textbackslash n\textbackslash n---\textbackslash n\textbackslash nHere is the transformed narrative in the specified format.\textbackslash n\textbackslash n\textbackslash n\textbackslash n---\textbackslash n\textbackslash nHere is the transformed narrative in the format you requested.\textbackslash n\textbackslash n\textbackslash n\textbackslash n---\textbackslash n\textbackslash nHere is the transformed narrative in the specified format.\textbackslash n\textbackslash n\textbackslash n\textbackslash n---\textbackslash n\textbackslash nHere is the transformed narrative in the format you requested. ... \textit{(omitted)}''
} \\
\midrule
\makecell[{{p{2cm}}}]{
  Missing \\ Required \\ Components
} &
\makecell[{{p{13cm}}}]{
  ``Explanation\textbackslash n\textbackslash n1. To solve this task it's best to think in terms of a small internal state we carry around, which changes depending on which operation we're considering.\textbackslash n2. Initially, all 1s are at the beginning of the string and all 0s at the end, so we can choose the entire string and get the answer we desire.\textbackslash n3. After the first operation, we would get the string 01111011, but if we look at the answer state we had previously, we see that we had 1110100 at the end, which is shorter. So we revert the last operation and pick the next substring of 01111, and repeat the process.\textbackslash n4. After the second operation, the answer state would become 11000010, which is still worse than our previous internal state. So we go back one more step and choose the entire string again, finally resulting in 010110.\textbackslash n5. Applying the steps to the other test cases follows a very similar logic, making use of the internal state to make decisions and progress.''
} \\
\bottomrule
\end{tabular}
\end{table*}

\section{The use of LLMs}

We used LLMs only for minor language editing, including adjustments to word choices and clarity. LLMs were not used for the research design, analysis, interpretation, or manuscript preparation.

\end{document}